%% file: ICLR_27/main.tex
\documentclass{article} 
\usepackage{ICLR_27/iclr2027_conference,times}

\input{ICLR_27/math_commands.tex}

\usepackage{url}
\usepackage{tikz}
\usepackage{wrapfig}

\title{PCAsplat: Gaussian Splatting with Local PCA Regularization}

\author{Vitor Matias$^{1}$\quad 
  Filipe Nascimento$^{1}$\quad 
  Kiyohiro Nakayama$^{2}$\quad 
  Jo{\~a}o Paulo Lima$^{3}$\quad \\
  \textbf{M{\'a}rcus Lobo}$^{1}$\quad 
  \textbf{Gordon Wetzstein}$^{2}$\quad 
  \textbf{Leonidas Guibas}$^{2}$\quad
  \textbf{Afonso Paiva}$^{1}$\quad
  \textbf{Tiago Novello}$^{2,4}$ \\[1ex]
  $^{1}$Universidade de S{\~a}o Paulo \quad
  $^{2}$Stanford University \quad
  $^{3}$UFRPE \quad
  $^{4}$IMPA \\[1ex]
}

\usepackage{listings}
\definecolor{codegreen}{rgb}{0,0.6,0}
\definecolor{codegray}{rgb}{0.5,0.5,0.5}
\definecolor{codepurple}{rgb}{0.58,0,0.82}
\definecolor{backcolour}{rgb}{0.96,0.96,0.94}

\lstdefinestyle{siggraphstyle}{
    backgroundcolor=\color{backcolour},   
    commentstyle=\color{codegreen},
    keywordstyle=\color{magenta},
    numberstyle=\tiny\color{codegray},
    stringstyle=\color{codepurple},
    basicstyle=\ttfamily\footnotesize,
    breakatwhitespace=false,         
    breaklines=true,                 
    captionpos=b,                    
    keepspaces=true,                 
    numbers=left,                    
    numbersep=5pt,                  
    showspaces=false,                
    showstringspaces=false,
    showtabs=false,                  
    tabsize=4
}
\usepackage{graphicx}
\usepackage{booktabs}
\usepackage{xcolor}
\definecolor{sigblue}{rgb}{0.21,0.49,0.74}

\newcommand{\method}{{PCAsplat}}
\usepackage{multirow}
\usepackage{threeparttable} 

\usepackage[normalem]{ulem}
\useunder{\uline}{\ul}{}
\iclrfinalcopy 

\usepackage{float}

\usepackage{hyperref}
\definecolor{sigblue}{rgb}{0.21,0.49,0.84}
\hypersetup{
colorlinks=true,
linkcolor=sigblue,
citecolor=sigblue,
urlcolor=sigblue,
filecolor=sigblue
}%

\usepackage[skip=4pt]{caption}                      

\begin{document}

\maketitle

\begin{abstract}
Gaussian splatting has emerged as a flexible representation for 3D reconstruction from posed images. However, existing methods are optimized primarily using rasterization-based losses, which supervise a splat only when it contributes to sampled camera rays.
Gaussians that are occluded or contribute little to the sampled view therefore receive weak or no geometric gradients and may drift away from the underlying surface, producing undesired floaters.
%
We introduce \method{}, a geometry-aware regularization framework for Gaussian splatting based on differentiable local principal component analysis (PCA). Our PCA regularizer acts directly on neighborhoods of Gaussian centers and can therefore update Gaussians that do not contribute to the current training view. We regularize the PCA eigenvalues to encourage Gaussians to move to the underlying surface with isotropic tangent-plane coverage. We also align each Gaussian normal with the PCA-estimated neighborhood normal to enforce consistent orientation. 
%
Experiments on DTU, Tanks and Temples, and NeRF Synthetic show that the splats produced by \method{} better approximate samples of the reference surface while substantially reducing undesired floaters. These surface-aligned splats enable downstream geometry-processing tasks, including point cloud segmentation, and direct Poisson reconstruction.  
Additionally, \method{} remains competitive under conventional novel view synthesis and mesh extraction tasks. Code will be released.

\end{abstract}

\input{sections/01_intro}
\input{sections/02_related_works}
\input{sections/03_background}
\input{sections/04_method}
\input{sections/05_results}

\input{sections/06_conclusion}



\subsection*{Acknowledgments}
This work was supported by the São Paulo Research~Foundation (FAPESP),  \#2026/01721-3 and \#2024/09462-1; Conselho Nacional de Desenvolvimento Científico e Tecnológico (CNPq) funding \#315158/2023-9; Google.

\bibliography{ICLR_27/bib}
\bibliographystyle{ICLR_27/iclr2027_conference}

\appendix
\input{sections/supp}

\end{document}

%% file: ICLR_27/math_commands.tex
\usepackage{amsmath,amsfonts,bm}

\def\eqref#1{equation~\ref{#1}}

\def\1{\bm{1}}

\DeclareMathAlphabet{\mathsfit}{\encodingdefault}{\sfdefault}{m}{sl}
\SetMathAlphabet{\mathsfit}{bold}{\encodingdefault}{\sfdefault}{bx}{n}



%% file: sections/01_intro.tex
\section{Introduction}

\begin{figure}[t]
\centering
  \includegraphics[width=\textwidth]{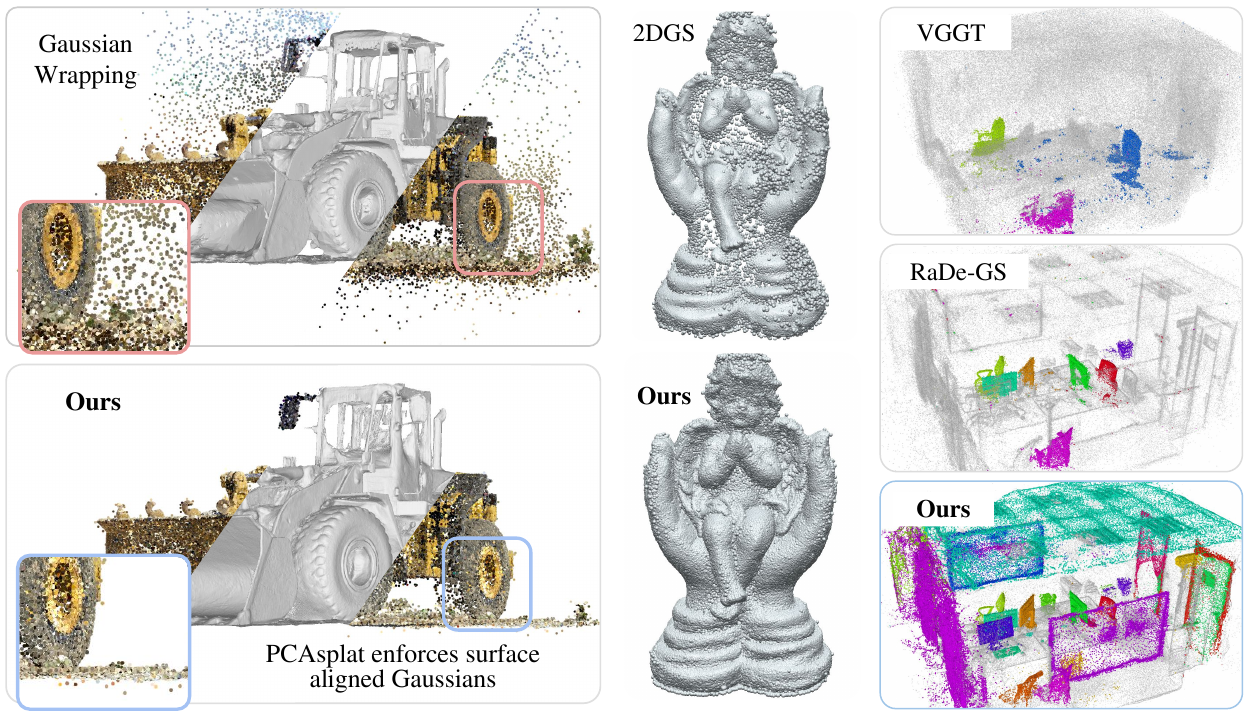}
  \caption{We present \textbf{\method{}}, a geometric regularization of Gaussian splatting that aligns Gaussians with surfaces through differentiable local PCA.
Compared with current methods, \method{} substantially reduces floaters (left) and produces cleaner point clouds with better surface coverage (center).
This improved point-cloud enables downstream tasks such as instance segmentation (right).}
  \label{fig:teaser}
\end{figure}

Gaussian splatting~\citep{kerbl20233d} is becoming the standard framework for novel view synthesis (NVS) and 3D reconstruction from images, offering a favorable trade-off between rendering quality, optimization speed, and interactive visualization.
It represents a scene as a collection of anisotropic Gaussians with appearance attributes optimized via differentiable rendering.
While Gaussian splatting is highly effective for NVS, Gaussians can be located far from the underlying surfaces, forming \textit{floaters} both outside and inside objects (see \autoref{fig:teaser}, top-left). These floaters remain a challenge for using the learned representation directly in geometry-processing tasks such as segmentation and Poisson surface reconstruction. %

Methods such as 2DGS~\citep{huang20242d}, GOF~\citep{yu2024gaussian}, PGSR~\citep{chen2024pgsr}, and RaDe-GS~\citep{zhang2026rade} use losses based on rendered depth and normals to improve geometric reconstruction.
However, these rendering-based losses supervise Gaussians through their contributions to sampled views.
Gaussians that are occluded or contribute little to the view therefore receive weak or no geometric gradients from these losses and may drift away from the underlying surface, producing exterior and interior floaters. The learned representation may also contain oversized, diffuse, or poorly localized Gaussians. Although floaters may have little impact on training views, exterior ones can appear in novel views while interior ones may remain occluded, with both limiting the representation's use in downstream applications.
Other approaches couple Gaussians to an auxiliary surface representation. GS-Pull~\citep{zhang2024gspull} pulls Gaussians onto the zero-level set of a learned neural signed distance function (SDF), while MILo~\citep{guedon2025milo} differentiably extracts a mesh from Gaussian parameters during training and enforces consistency between the mesh and the Gaussians. However, these approaches rely on an auxiliary representation that must also be optimized, and undesired floaters remain an issue.

Overall, learning surface-aligned Gaussians suitable for downstream geometry processing remains a challenge. To address this problem, we introduce \method{}, a geometry-aware regularization framework for Gaussian splatting based on differentiable local principal component analysis (PCA).
Our key idea is to regularize Gaussian neighborhoods directly in scene space through PCA statistics. This provides geometric supervision even to Gaussians that do not contribute to the current training view.
For each Gaussian, we estimate a local 3D neighborhood and compute its covariance eigenspectrum to characterize local thickness and tangent-plane anisotropy. We use these statistics to encourage Gaussian centers to move toward the surface by reducing local thickness, while promoting isotropic tangent-plane coverage and aligning Gaussian normals with PCA-derived normals up to sign.

Experiments on DTU, Tanks and Temples, and NeRF Synthetic show that \method{} achieves strong surface reconstruction results compared to current methods. It produces surface-aligned splats with fewer exterior and interior floaters (\autoref{fig:teaser}, bottom-left and center), allowing for more accurate downstream geometry processing. \method{} yields substantially more accurate instance segmentation and more coherent object masks than current methods using the same segmentation pipeline (\autoref{fig:teaser}, right). \method{} also remains competitive under conventional novel view synthesis and TSDF-based mesh extraction.
In summary, our \textbf{main contributions} are:
\begin{itemize}
    \item A differentiable local PCA framework for regularizing Gaussian neighborhoods directly in scene space, including Gaussians that do not contribute to the current training view.
    \item Geometric losses for learning surface-aligned Gaussians by moving them toward the surface, promoting isotropic coverage, and aligning their normals with PCA-normals.
    \item An evaluation of splats as surface samples, showing improved geometric accuracy and their use in downstream applications, including segmentation and Poisson reconstruction.
\end{itemize}

%


%% file: sections/02_related_works.tex
\section{Related Work}


\textbf{Surface reconstruction from images} is a long-standing problem in vision. Classical methods such as COLMAP~\citep{schoenberger2016sfm} combine structure from motion and multi-view stereo to recover camera poses and point clouds. Recently, feed-forward models such as VGGT~\citep{wang2025vggt} and VGGT-$\Omega$~\citep{wang2026vggtomega} predict cameras, depth maps, and point clouds directly from unposed images. However, these methods lack a view-dependent appearance model for novel-view synthesis.
In contrast, neural implicit methods jointly learn geometry and view-dependent appearance: VolSDF~\citep{yariv2021volume}, NeuS~\citep{wang2021neus}, and Neuralangelo~\citep{li2023neuralangelo} optimize neural SDFs for each scene through differentiable rendering. However, repeated neural-field evaluations along each ray make volume rendering computationally expensive. Gaussian splatting~\citep{kerbl20233d} enabled fast rasterization of explicit primitives but lacked geometry constraints. 2DGS~\citep{huang20242d}, Gaussian Surfels~\citep{dai2024gaussiansurfels}, and Quadratic Gaussian Splatting~\citep{zhang2025quadraticgaussiansplattinghigh} made splats surface-like through modified primitives or rendering formulations together with rendering losses to regularize geometric properties of the scene. SuGaR~\citep{guedon2024sugar}, GOF~\citep{yu2024gaussian}, and RaDe-GS~\citep{zhang2026rade} improve mesh extraction via surface alignment, opacity fields, or better depth rasterization. And, PGSR~\citep{chen2024pgsr} introduced multi-view consistency losses to further improve geometry.
For mesh extraction, methods such as 2DGS, RaDe-GS, and PGSR fuse rendered depth maps using TSDF~\citep{curless1996volumetric}, which are based on ray-gaussian contributions. Consequently, accurate meshes do not necessarily imply surface-aligned Gaussian centers.

\textbf{Rendering-based geometric regularization} augments photometric supervision with losses on rendered depth and normals, aiming to reduce undesired \textbf{floaters}. Mip-NeRF~360~\citep{barron2022mip} introduces a distortion loss that encourages rendering contributions to concentrate within a narrow interval along each ray, discouraging diffuse density and floaters. 2DGS and GOF adapt this loss to encourage surface alignment by reducing depth differences between contributing Gaussians along viewing rays. These differences are weighted by rendering contributions, which depend on opacity and accumulated transmittance. 
As such, Gaussians that are occluded or contribute little to a view may receive weak or no geometric gradients from these losses, allowing exterior and interior floaters to persist. \method{} complements this supervision with local PCA losses that act directly on 3D Gaussian neighborhoods, so Gaussians that do not contribute to the current training view can still receive geometric gradients.

\textbf{Hybrid Gaussian--surface representations} couple Gaussians with auxiliary surface representations. GS-Pull~\citep{zhang2024gspull} pulls Gaussian centers onto the zero-level set of a neural SDF, at the additional computational cost of jointly optimizing this field. Its Gaussian-disk supervision addresses sparse, nonuniform support but remains sensitive to extreme sparsity. GSDF~\citep{yu2024gsdf} couples Gaussians and SDFs through depth and normal consistency, using Gaussian-rendered depth to guide SDF sampling and SDF values to control Gaussian growth and pruning. Gaussian-UDF~\citep{li2025gaussianudf} jointly learns 2D Gaussians and an unsigned distance field for open surfaces, using Gaussian centers and local planes to supervise distances. MILo~\citep{guedon2025milo} and GaussianWrapping (GW)~\citep{gomez2026blobsspokeshighfidelitysurface} differentiably extracts a mesh from Gaussian-derived pivots and learned signed distances, enforcing Gaussian--mesh rendering consistency.

In contrast, we adapt local PCA, a classical method for estimating tangent planes and normals from point clouds~\citep{hoppe1992surface,pauly2002efficient}, to directly regularize Gaussian neighborhoods. We add losses for neighborhood thickness, tangent-plane coverage, and normal alignment to existing splatting objectives without learning auxiliary SDFs or extracting meshes during training.

%% file: sections/04_method.tex
\section{Method}

\begin{figure}[h!]
    \centering
    \includegraphics[width=\textwidth]{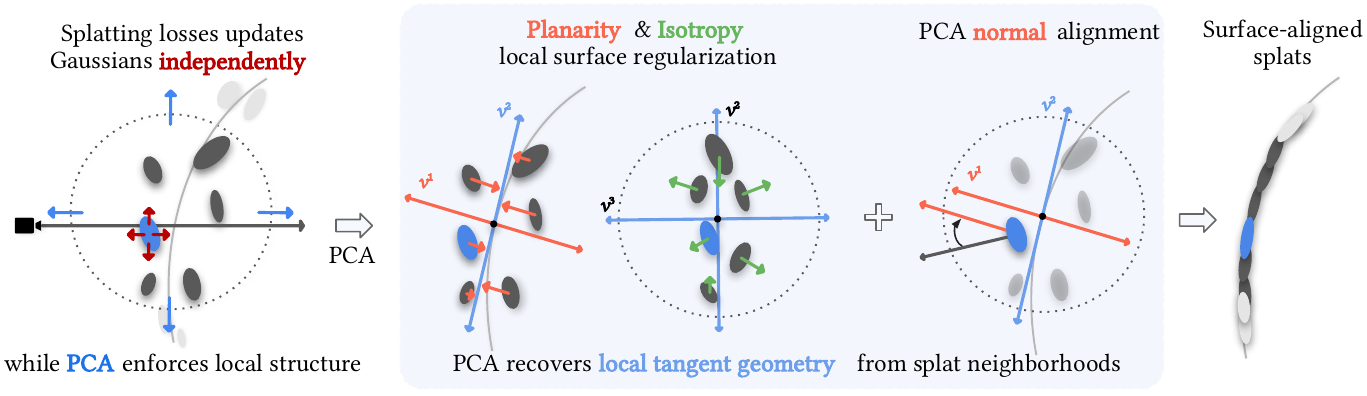}
    \caption{
    \textbf{Overview of \method.}
    We regularize Gaussian neighborhoods directly in scene space.
    For each Gaussian, we estimate a local neighborhood and compute differentiable PCA statistics that capture the underlying tangent structure.
    We then regularize the local eigenspectrum to reduce neighborhood thickness and promote isotropic tangent-plane coverage. We also align Gaussian normals with PCA-derived normals, encouraging surface-aligned and well-localized splats.
    Our formulation is lightweight, differentiable, and directly compatible with existing splatting pipelines.}
    \label{fig:overview}
 \end{figure}

Our goal is to learn surface-aligned Gaussians by regularizing their local geometry directly in scene space, reducing floaters and supporting downstream geometry processing. We formulate local PCA-based regularization as differentiable losses on Gaussian geometry, optimized jointly with the rendering objective. For each Gaussian, we compute the covariance eigenspectrum and principal directions of a local neighborhood of Gaussian centers. These quantities define losses that reduce neighborhood thickness, enforce isotropic tangent-plane coverage, and align Gaussian normals with PCA-derived normals. \autoref{fig:overview} gives an overview of our framework.

\subsection{Problem setting}

Given a set of posed images $\{I_i\}_{i=1}^{M}$, our goal is to optimize a Gaussian splatting point cloud to be geometrically aligned with the underlying scene surface. Recall that Gaussian splatting represents a scene as a set of anisotropic Gaussians
$\mathcal{G}=\{g_k\}_{k=1}^N$,
$g_k = (\mu_k, \Sigma_k, \alpha_k, c_k),
$
where $\mu_k \in \mathbb{R}^3$ denotes the Gaussian center, $\Sigma_k \in \mathbb{R}^{3\times3}$ its positive-definite covariance matrix, $\alpha_k\in[0,1]$ its opacity, and $c_k$ its view-dependent RGB color, represented using spherical harmonics.
Each Gaussian defines kernels
$
g_k(x)
=
\exp\!\left(
-\frac{1}{2}\left\langle x-\mu_k,\Sigma_k^{-1}(x-\mu_k)\right\rangle
\right).
$
which are projected onto the image plane and rasterized through alpha compositing. For a view ray $r$, with Gaussians indexed in front-to-back order, the rendered color is given by
\[
\hat I(r)
=
\sum_{k=1}^{N}
T_k(r)\,\alpha_k(r)\,c_k
+T_{N+1}(r)\,c_{\mathrm{bg}},
\qquad
T_k(r)
=
\prod_{j=1}^{k-1}\bigl(1-\alpha_j(r)\bigr),
\]
where $\alpha_k(r)\!\!=\!\!\alpha_kg_k^{2\mathrm{D}}(r)$ is the effective opacity, with $g_k^{2\mathrm{D}}(r)$ denoting the projected Gaussian evaluated at the pixel corresponding to $r$. Here, $T_k(r)$ denotes the transmittance, $c_k$ is evaluated for the current view, and $c_{\mathrm{bg}}$ is the background color.
$\mathcal{G}$ is optimized using the standard photometric loss
\begin{align}\label{e-rgb}
\mathcal{L}_{\mathrm{RGB}}
=
(1-\lambda)\,
\|I-\hat I\|_1
+
\lambda\,
\mathcal{L}_{\mathrm{SSIM}}(I,\hat I),
\end{align}
where $I$ and $\hat I$ denote the ground-truth and rendered images, $\lambda\in[0,1]$ balances the two terms, and $\mathcal{L}_{\mathrm{SSIM}}(I,\hat I)=1-\operatorname{SSIM}(I,\hat I)$ denotes the structural dissimilarity loss. The $\ell_1$ term is averaged over pixels and RGB channels. While Gaussian splatting is effective for appearance reconstruction, this photometric objective does not explicitly regularize the local geometry of Gaussian neighborhoods.


We therefore use differentiable local PCA (Section~\ref{s-local_PCA}) to regularize Gaussian neighborhoods directly in scene space. Our losses reduce neighborhood thickness, enforce isotropic tangent-plane coverage, and align Gaussian normals with PCA-derived normals (Section~\ref{s-PCA_losses}).


\subsection{Local PCA neighborhoods}
\label{s-local_PCA}

For each Gaussian $g_i$, we define its local neighborhood $\mathcal{N}_K(i) = \mathrm{KNN}(\mu_i, K)$ as the indices of the $K$ nearest Gaussian centers to $\mu_i\in \mathbb{R}^3$.
As in classical point-cloud normal estimation~\citep{hoppe1992surface,pauly2002efficient}, we compute the covariance matrix of the neighboring centers:
\begin{align}\label{e-covariance_matrix}
\mathbf{C}_i = \frac{1}{K}\sum_{j \in \mathcal{N}_K(i)}
(\mu_j - \bar{\mu}_i)(\mu_j - \bar{\mu}_i)^\top,
\qquad
\bar{\mu}_i = \frac{1}{K}\sum_{j \in \mathcal{N}_K(i)} \mu_j.
\end{align}
Here, $\bar{\mu}_i$ is the neighborhood centroid.
The eigendecomposition of $\mathbf{C}_i$,
$\mathbf{C}_i \mathbf{v}_i^{(m)} = \lambda_i^{(m)} \mathbf{v}_i^{(m)}$, with orthonormal eigenvectors and
$0 \le \lambda_i^{(1)} \le \lambda_i^{(2)} \le \lambda_i^{(3)}$,
summarizes the local geometric structure. The eigenvector $\mathbf{v}_i^{(1)}$ associated with the smallest eigenvalue estimates the local surface \textit{normal}, while $\mathbf{v}_i^{(2)}$ and $\mathbf{v}_i^{(3)}$ span the estimated \textit{tangent plane}.
The plane through $\bar\mu_i$ with these tangent directions locally approximates the underlying surface. Since $\lambda_i^{(1)}$ measures the mean squared distance of neighboring Gaussian centers to this plane, we minimize it to pull the centers toward the plane during optimization. Optimizing this objective jointly with the rendering loss requires computing its gradient with respect to the Gaussian centers.

Assuming fixed neighborhood indices and no repeated eigenvalues, the partial derivatives needed for backpropagation are~\citep{greenbaum2020first}
\[
\frac{\partial\lambda_i^{(m)}}{\partial\mu_j}
=\frac{2}{K}\left\langle\mathbf{v}_i^{(m)},\mu_j-\bar\mu_i\right\rangle\mathbf{v}_i^{(m)},
\qquad j\in\mathcal N_K(i).
\]
The gradient is proportional to the projection of $\mu_j-\bar\mu_i$ onto $\mathbf v_i^{(m)}$; for $m=1$, its negative points toward the estimated tangent plane. We use these derivatives to optimize the eigenspectrum loss introduced next directly with respect to the Gaussian centers.

\subsection{PCA-based geometric losses}
\label{s-PCA_losses}

\paragraph{PCA eigenspectrum loss.}
We use local PCA to move the Gaussian centers $\{\mu_i\}_{i=1}^N$ toward the underlying surface.
In other words, surface-like Gaussian neighborhoods should be thin along the PCA normal direction and well distributed within the tangent plane of the underlying surface, i.e.,
$
\lambda_i^{(1)} \ll \lambda_i^{(2)} \approx \lambda_i^{(3)}.
$
We therefore regularize the local PCA eigenspectrum using
\begin{equation}
\mathcal{L}_{\mathrm{PCA}} = \frac{1}{N}\sum_{i=1}^{N}
\Bigl[
\lambda_i^{(1)} + \gamma\,\bigl|\lambda_i^{(3)}-\lambda_i^{(2)}\bigr|
\Bigr].
\end{equation}

\begin{wrapfigure}[11]{r}{0.35\textwidth}
    \centering
    \vspace{-0.45cm}
    \begin{tikzpicture}
        \node[anchor=south west, inner sep=0, outer sep=0]
            (img) at (0,0)
            {\includegraphics[width=\linewidth]{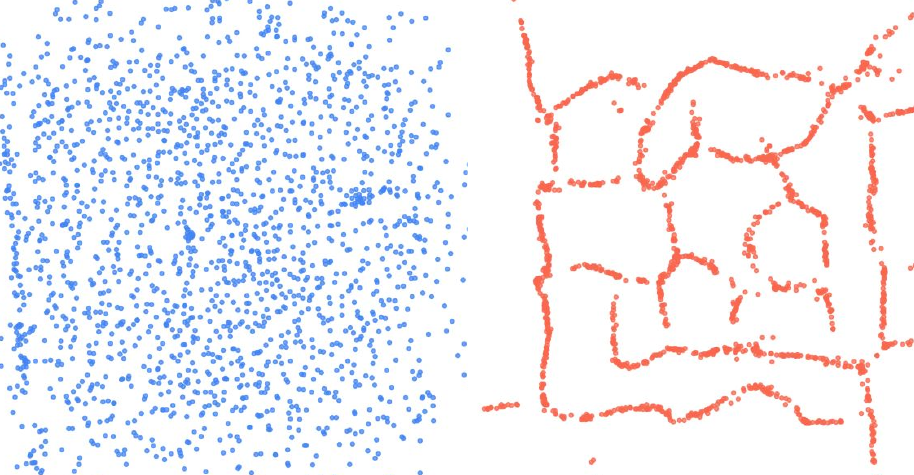}};

        \begin{scope}[
            x={(img.south east)},
            y={(img.north west)},
            every node/.style={anchor=base, font=\small, yshift=3pt}
        ]
            \node at (0.25, 1.08) {\footnotesize With $\left| \lambda_i^{(3)}-\lambda_i^{(2)} \right|$};
            \node at (0.75, 1.08) {\footnotesize Without};
        \end{scope}
    \end{tikzpicture}
    \caption{Balanced coverage with isotropy (left); elongated structures without it (right).}
    \label{fig:pca_isotropy}
\end{wrapfigure}
The first term penalizes variance along the estimated normal, while the second promotes isotropic tangent-plane coverage. We balance these terms with $\gamma=0.001$.

This loss acts directly on the spatial distribution of Gaussian centers, rather than on rendered depth or normal maps. Minimizing $\lambda_i^{(1)}$ pulls neighboring Gaussian centers toward the estimated tangent plane through $\bar\mu_i$, discouraging floaters. The tangent isotropy term discourages elongated neighborhoods by encouraging balanced spread along the two tangent directions, as shown in \autoref{fig:pca_isotropy}. Together, these terms encourage thin neighborhoods with balanced tangent-plane support, improving the samples used for downstream mesh extraction and point-cloud processing.

\paragraph{PCA normal alignment.}
To encourage Gaussians to align with the local geometry, we define a normal consistency loss between each Gaussian normal and the local PCA normal.
Specifically, each Gaussian $g_i$ has a rotation matrix $\mathbf{R}_i \in \mathrm{SO}(3)$, with $\mathbf{R}_i = (\mathbf{r}_{i,1},\mathbf{r}_{i,2},\mathbf{r}_{i,3})$, and a scaling vector $\mathbf{s}_i = (s_{i,1},s_{i,2},s_{i,3}) \in \mathbb{R}_{>0}^3$. These parameters define its covariance as $\Sigma_i=\mathbf R_i\operatorname{diag}(s_{i,1}^2,s_{i,2}^2,s_{i,3}^2)\mathbf R_i^\top$.
We define the Gaussian normal $\mathbf{n}_i^g$ as the axis associated with the smallest scale:
\[
\mathbf{n}_i^g = \mathbf{r}_{i,k^*},
\qquad
k^* = \arg\min_{1\leq k \leq 3} s_{i,k}.
\]
We combine sign-invariant alignment between the PCA normal $\mathbf{v}_i^{(1)}$ and the Gaussian normal $\mathbf{n}_i^g$ with a scale penalty in the normal loss
\begin{equation}\label{e-normal}
\mathcal{L}_{\mathrm{normal}}
=
w_{\mathrm{normal}}\, \frac{1}{N}
\sum_{i=1}^{N}
\left(
1 - \left|\left\langle\mathbf{v}_i^{(1)},\mathbf{n}_i^g\right\rangle\right|\right)
\,+w_{\mathrm{surfel}}\,
\frac{1}{N}
\sum_{i=1}^{N}
 \min(s_{i,1},s_{i,2},s_{i,3}).
\end{equation}
The second term encourages each Gaussian to behave as a 2D surfel by penalizing its smallest scale, with $w_{\mathrm{normal}}$ and $w_{\mathrm{surfel}}$ controlling the strength of this regularization.

\paragraph{Final loss function.}
Together, the PCA eigenspectrum and normal alignment terms encourage Gaussian centers and orientations to form coherent, disk-like local surface structures. Thus, our core training objective is
\begin{equation}
\mathcal{L}
=
\mathcal{L}_{\mathrm{RGB}}
+
w_{\mathrm{PCA}}\mathcal{L}_{\mathrm{PCA}}
+
\mathcal{L}_{\mathrm{normal}},
\end{equation}
where $\mathcal{L}_{\mathrm{RGB}}$ is the photometric rendering loss in \autoref{e-rgb}, $\mathcal{L}_{\mathrm{PCA}}$ regularizes the local eigenspectrum with $w_{\mathrm{PCA}}$ being scheduled to start at 1 and grow to 10, and $\mathcal{L}_{\mathrm{normal}}$ aligns Gaussian normals with PCA-derived normals. 
In our implementation, we also keep the standard PGSR multi-view loss $\mathcal{L}_{\mathrm{pgsr}}=\mathcal{L}_{\mathrm{mvgeo}}+\mathcal{L}_{\mathrm{ncc}}$ from the base pipeline, where $\mathcal{L}_{\mathrm{mvgeo}}$ enforces multi-view geometric consistency and $\mathcal{L}_{\mathrm{ncc}}$ measures patch-wise normal consistency across views. 


%% file: sections/05_results.tex
\section{Experiments}

We evaluate \method{} on the surface alignment of its learned Gaussians and their performance in downstream geometry processing. We compare against Gaussian-based reconstruction methods on point-cloud accuracy, 3D instance segmentation~\citep{schult2023mask3d}, and Poisson surface reconstruction~\citep{Kazhdanspsrpoisson}.

\textbf{Datasets and metrics.}
We evaluate surface accuracy on DTU~\citep{jensen2014large} and Tanks and Temples~\citep{Knapitsch2017tankstemples} using standard Chamfer distance (CD) and F1-score. Both metrics compare reconstructed point clouds (Gaussian centers for splatting methods) with ground-truth point clouds. For 3D semantic instance segmentation, we use ScanNet++~\citep{yeshwanthliu2023scannetpp} and report Top-1 and Top-3 mean intersection over union (mIoU). We also evaluate F1-score on a visible-surface benchmark constructed from the NeRF Synthetic~\citep{mildenhall2021nerf} scene assets. Implementation details are provided in \autoref{sec:implementation_details}, and the visible-surface reference meshes we introduce are described in \autoref{sec:nerf_syn_dataset}.

\textbf{Baselines.}
We compare with standard 3DGS and leading Gaussian splatting methods for surface reconstruction, including 2DGS~\citep{huang20242d}, PGSR~\citep{chen2024pgsr}, GS-Pull~\citep{zhang2024gspull}, MILo~\citep{guedon2025milo}, RaDe-GS~\citep{zhang2026rade}, and GaussianWrapping (GW)~\citep{gomez2026blobsspokeshighfidelitysurface}. The latter use depth distortion, geometric consistency, or surface constraints to improve reconstruction. VGGT-$\Omega$~\citep{wang2026vggtomega} and COLMAP-D~\citep{schoenberger2016sfm} provide additional point-cloud references; ``-D'' denotes dense output.



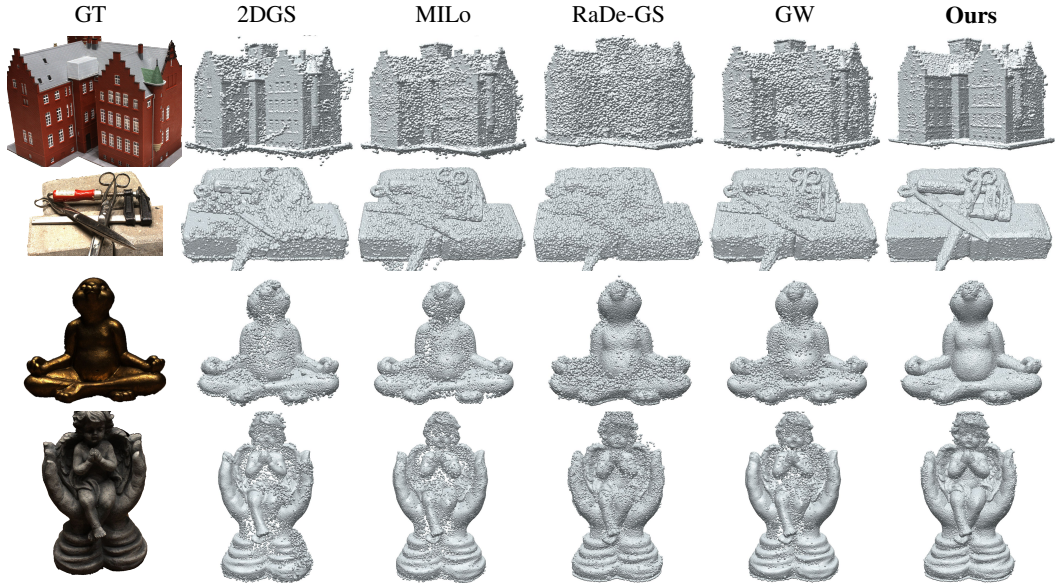
\begin{figure}[t]
    \centering
    \begin{tikzpicture}
        \node[anchor=south west, inner sep=0, outer sep=0]
            (img) at (0,0)
            {\includegraphics[width=\linewidth]{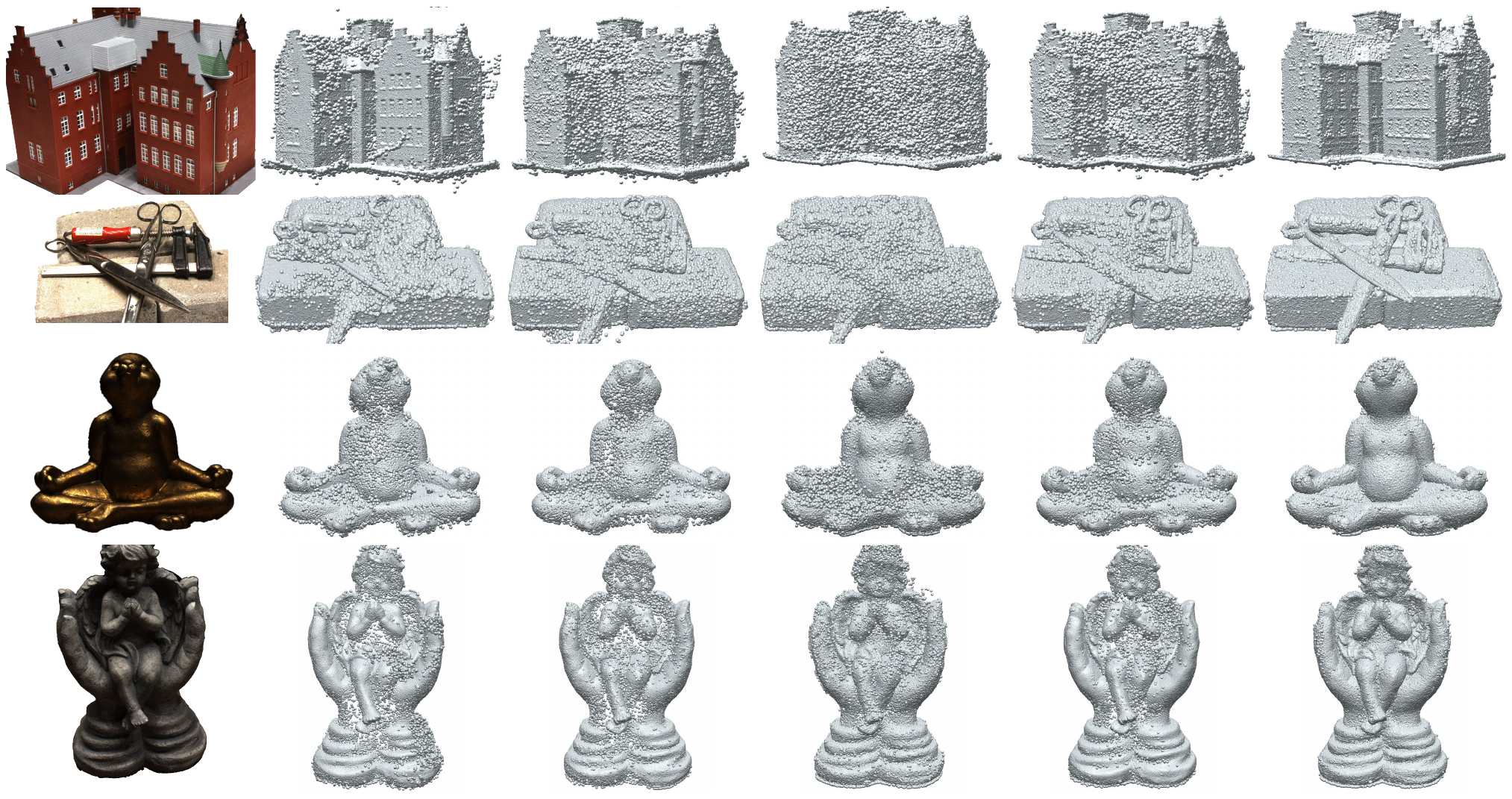}};

        \begin{scope}[
            x={(img.south east)},
            y={(img.north west)},
            every node/.style={anchor=north, font=\small}
        ]
            \node at (0.083, 1.061) {GT};
            \node at (0.250, 1.061) {2DGS};
            \node at (0.417, 1.061) {MILo};
            \node at (0.583, 1.061) {RaDe-GS};
            \node at (0.750, 1.061) {GW};
            \node at (0.917, 1.061) {\textbf{Ours}};
        \end{scope}
    \end{tikzpicture}

    \caption{\textbf{Qualitative comparison on DTU.} \method{} produces surface-aligned Gaussian centers with substantially fewer floaters, sharper edges, and finer geometric detail than the baselines.}
    \label{fig:pcd_align_dtu}
\end{figure}

\input{tabs/dtu_pcd_cd}

\begin{figure}[h]
    \centering
    \begin{tikzpicture}
    \node[anchor=south west, inner sep=0, outer sep=0]
        (img) at (0,0)
        {\includegraphics[width=\linewidth]{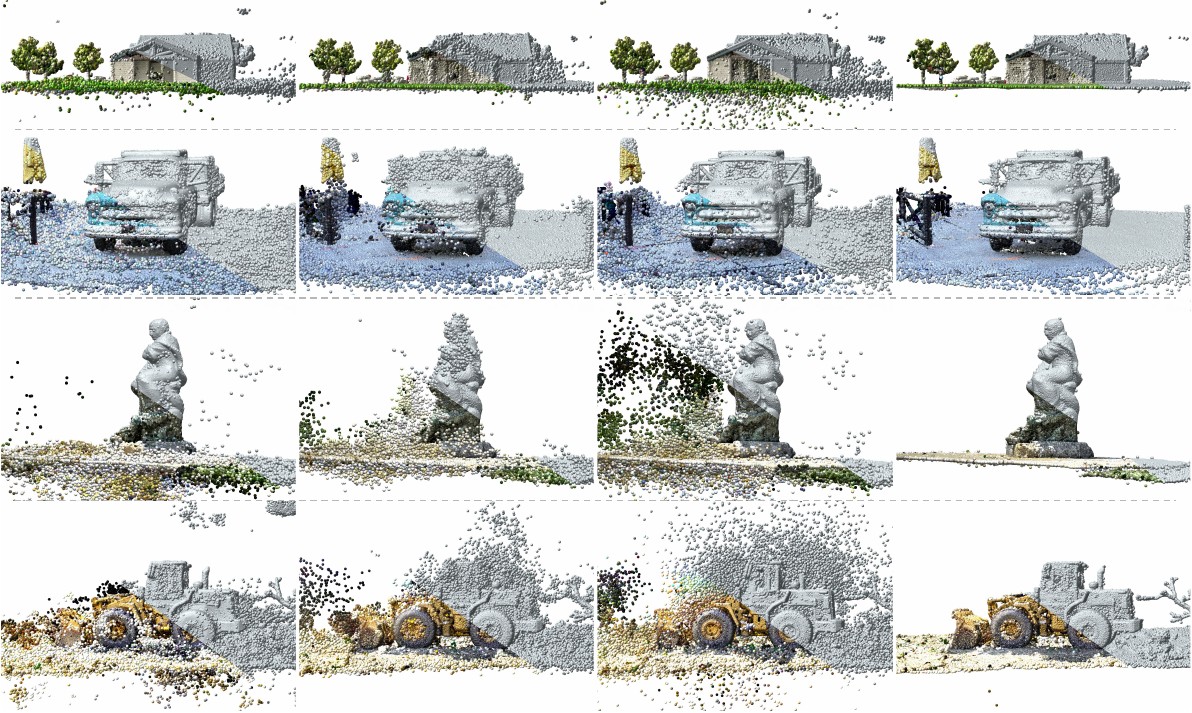}};

    \begin{scope}[
        x={(img.south east)},
        y={(img.north west)},
        every node/.style={anchor=north, font=\small}
    ]
        \node at (0.125,1.061) {MILo};
        \node at (0.375,1.061) {RaDe-GS};
        \node at (0.625,1.061) {GaussianWrapping};
        \node at (0.875,1.061) {\textbf{Ours}};
    \end{scope}
\end{tikzpicture}
    \vspace{-0.3cm}
    \caption{\textbf{Qualitative comparison on Tanks and Temples.}
    Compared with MILo, RaDe-GS, and GaussianWrapping, \method{} concentrates Gaussian centers along scene surfaces, with substantially fewer floaters around objects and below ground. See \autoref{fig:over_tnt} and \autoref{fig:under_tnt} for additional views of the geometry above and below ground.
    }
    \label{fig:pcd_align_tnt}
\end{figure}
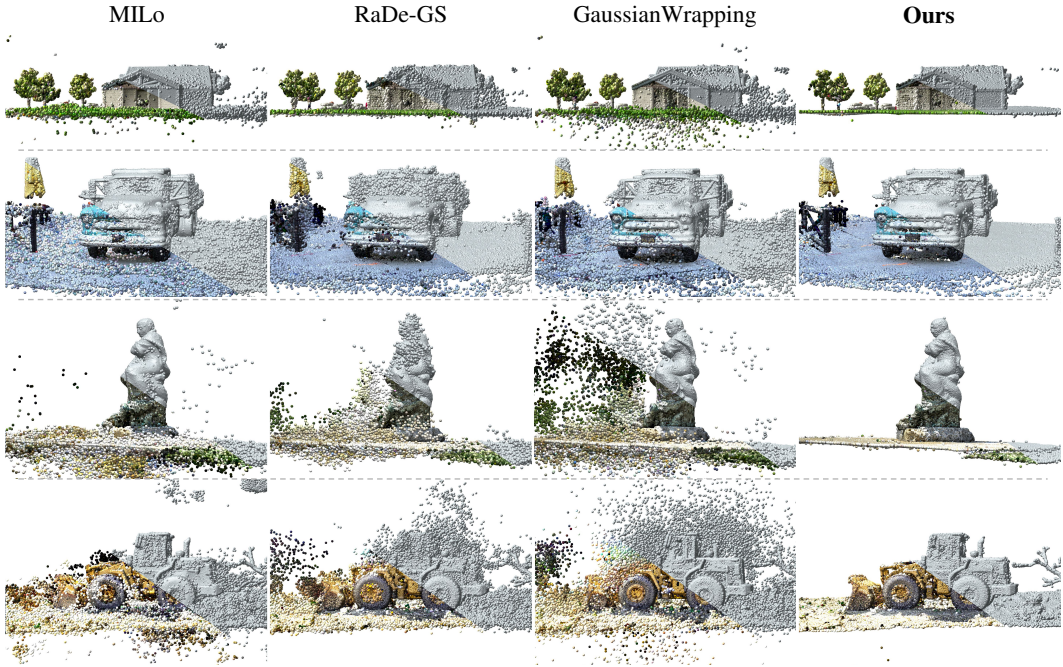

\begingroup
\setlength{\columnsep}{6pt}
\setlength{\intextsep}{2pt}
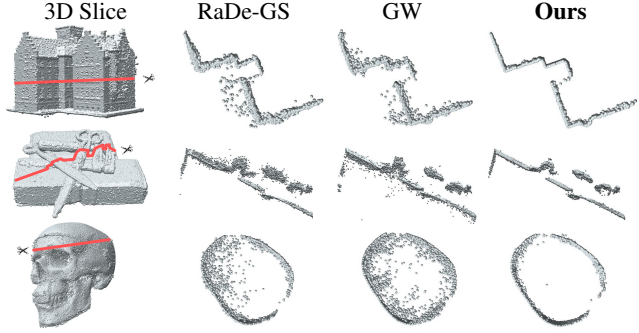
\begin{wrapfigure}[14]{r}{0.6\textwidth}
    \centering
    \vspace{-0.2cm}
    \begin{tikzpicture}
    \node[anchor=south west, inner sep=0, outer sep=0]
        (img) at (0,0)
        {\includegraphics[width=\linewidth]{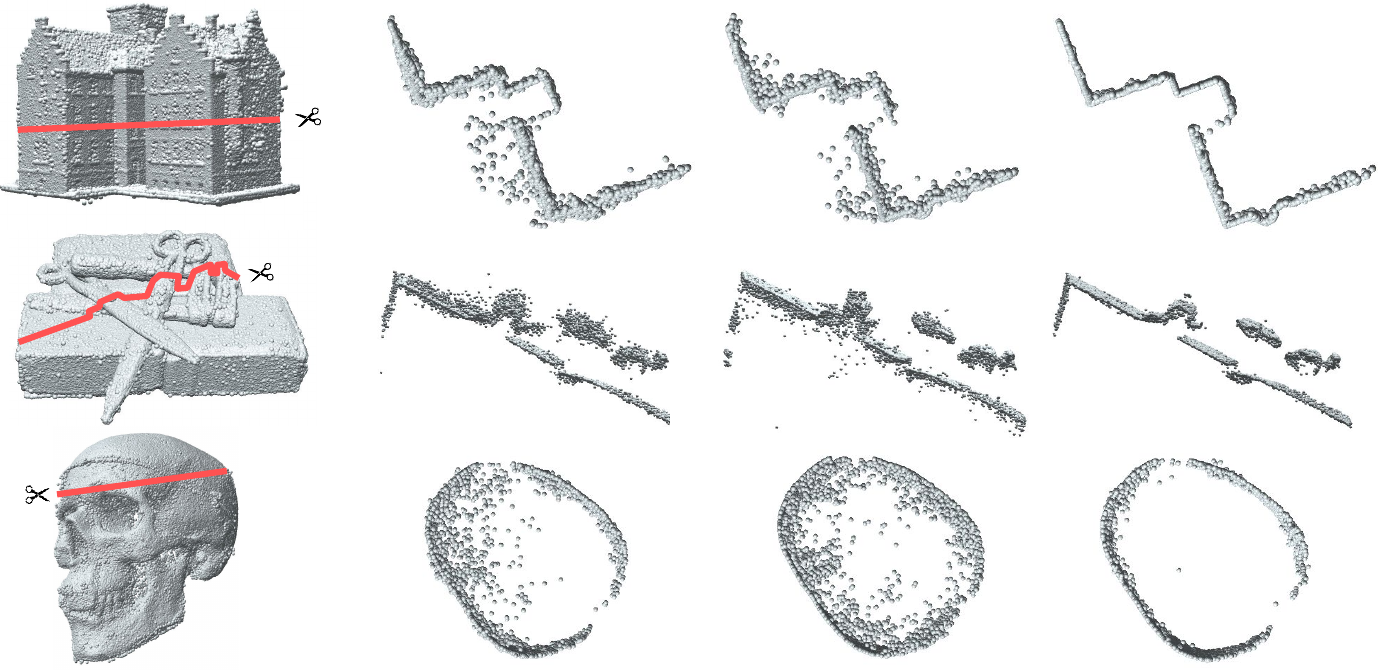}};

    \begin{scope}[
        x={(img.south east)},
        y={(img.north west)},
        every node/.style={anchor=north, font=\small}
    ]
        \node at (0.125,1.1) {3D Slice};
        \node at (0.375,1.1) {RaDe-GS};
        \node at (0.625,1.1) {GW};
        \node at (0.875,1.1) {\textbf{Ours}};
    \end{scope}
\end{tikzpicture}
    \caption{Qualitative comparison of cross-section slices on challenging DTU scenes.}
    \label{fig:slice_dtu_pcd}
\end{wrapfigure}

\subsection{Comparisons}
\autoref{tab:pcd_dtu} shows that \method{} outperforms all evaluated Gaussian-based methods in mean Chamfer distance on DTU. Our mean CD of $0.519$\,mm is $22\%$ lower than that of the strongest Gaussian-based baseline, RaDe-GS ($0.662$\,mm). While our method matches COLMAP-D's reported mean CD, we have a lower runtime ($22.3$ vs.\ $26.9$ minutes) and can perform novel view synthesis.
\autoref{fig:pcd_align_dtu} shows that \method{} produces surface-aligned Gaussian centers with sharper building edges, clearer tool boundaries, and finer statue details. The cross-sections in \autoref{fig:slice_dtu_pcd} show thinner surface profiles and substantially fewer \textbf{exterior and interior floaters} than RaDe-GS and GW. In the head cross-section, our Gaussian centers concentrate around the surface, as expected, while both baselines leave many centers scattered inside the object.

\begin{wraptable}[10]{r}{0.30\textwidth}
  \centering
  \input{tabs/dtu_plus_pca_cd}
\end{wraptable}
To show that our local PCA regularization benefits other splatting methods, we apply it to three representative baselines (2DGS, PGSR, and MILo) without additional hyperparameter tuning. All three achieve \textbf{lower Chamfer distance} (\autoref{tab:other_plus_pca}), showing that our regularization improves Gaussian geometry as a \textbf{plug-and-play} addition to existing methods.

On Tanks and Temples, \method{} achieves the highest mean F1-score among the evaluated Gaussian-based methods (\autoref{tab:tnt_nerf_f1_pcd}). The qualitative comparisons in \autoref{fig:pcd_align_tnt} show fewer floaters and better surface coverage than the Gaussian-based baselines shown. On the new NeRF Synthetic visible mesh benchmark, \method{} also achieves the highest mean F1-score among the compared methods and the best or second-best score on every scene. By using reference meshes derived directly from the original synthetic scene assets, this benchmark evaluates alignment to known surfaces and complements DTU and Tanks and Temples, whose captured reference geometry can contain inaccuracies~\citep{guedon2025milo,gomez2026blobsspokeshighfidelitysurface}. Together, these results show that \method{} improves Gaussian-to-surface alignment on both captured and synthetic scenes.


\autoref{ap:additional_results} expands our evaluation. First, our surface-aligned splats are asset-ready, allowing object extraction through spatial cropping without manual floater removal. Second, aerial and below-ground views of Tanks and Temples show Gaussian centers better aligned with the ground and walls, with fewer floaters. Finally, for TSDF-based mesh extraction and novel view synthesis, \method{} achieves the lowest mean Chamfer distance on DTU while remaining competitive on Tanks and Temples.

\par
\endgroup
\input{tabs/merged_tnt_nerf_f1_pcd}

\subsection{Ablation study}
\begin{wraptable}[6]{r}{0.24\textwidth}
  \centering
  \vspace{-0.3cm}
    \fontsize{8}{9}\selectfont
    \setlength{\tabcolsep}{2pt}
    \renewcommand{\arraystretch}{1}
    \begin{tabular}{@{}ccc@{}}
      \toprule
      $\mathcal{L}_{\mathrm{PCA}}$ & $\mathcal{L}_{\mathrm{normal}}$ & CD $\downarrow$ \\ \midrule
      & & 0.655 \\
      & \checkmark & 0.654 \\
      \checkmark & & 0.520 \\
      \checkmark & \checkmark & \textbf{0.519} \\ \bottomrule
    \end{tabular}
    \caption{\small\textbf{Ablation study}.}
    \label{tab:ablation}
\end{wraptable}
We evaluate the PCA loss $\mathcal{L}_{\mathrm{PCA}}$ and the normal alignment loss $\mathcal{L}_{\mathrm{normal}}$, individually and together. \autoref{tab:ablation} reports Chamfer distance between Gaussian centers and DTU ground-truth point clouds. The PCA loss, which penalizes neighborhood thickness and promotes isotropic tangent-plane coverage, accounts for most of the improvement, reducing CD from $0.655$ to $0.520$. Combining it with $\mathcal{L}_{\mathrm{normal}}$ yields the lowest CD of $0.519$.

\subsection{Applications}

We use the learned Gaussian centers directly as surface samples for instance segmentation and Poisson surface reconstruction. These experiments test how improved Gaussian-to-surface alignment benefits downstream geometry processing.

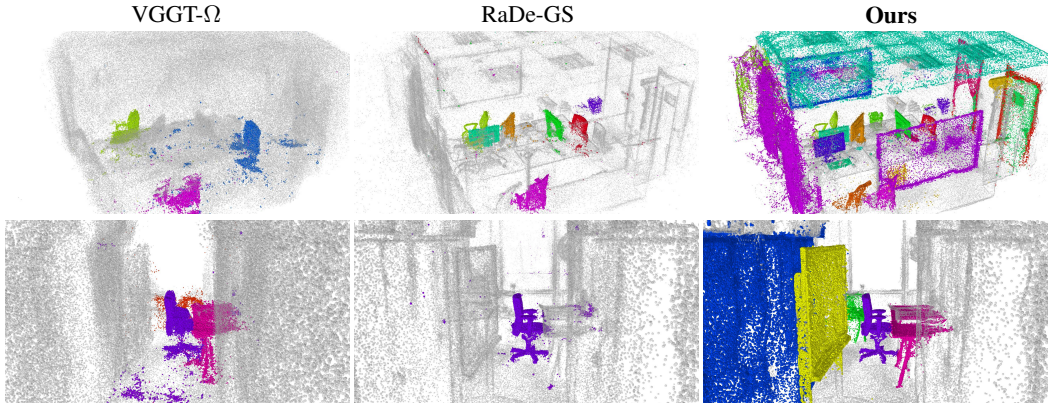
\begin{figure}[h]
    \centering

    \begin{tikzpicture}
        \node[anchor=south west, inner sep=0, outer sep=0]
            (img) at (0,0)
            {\includegraphics[width=\linewidth]{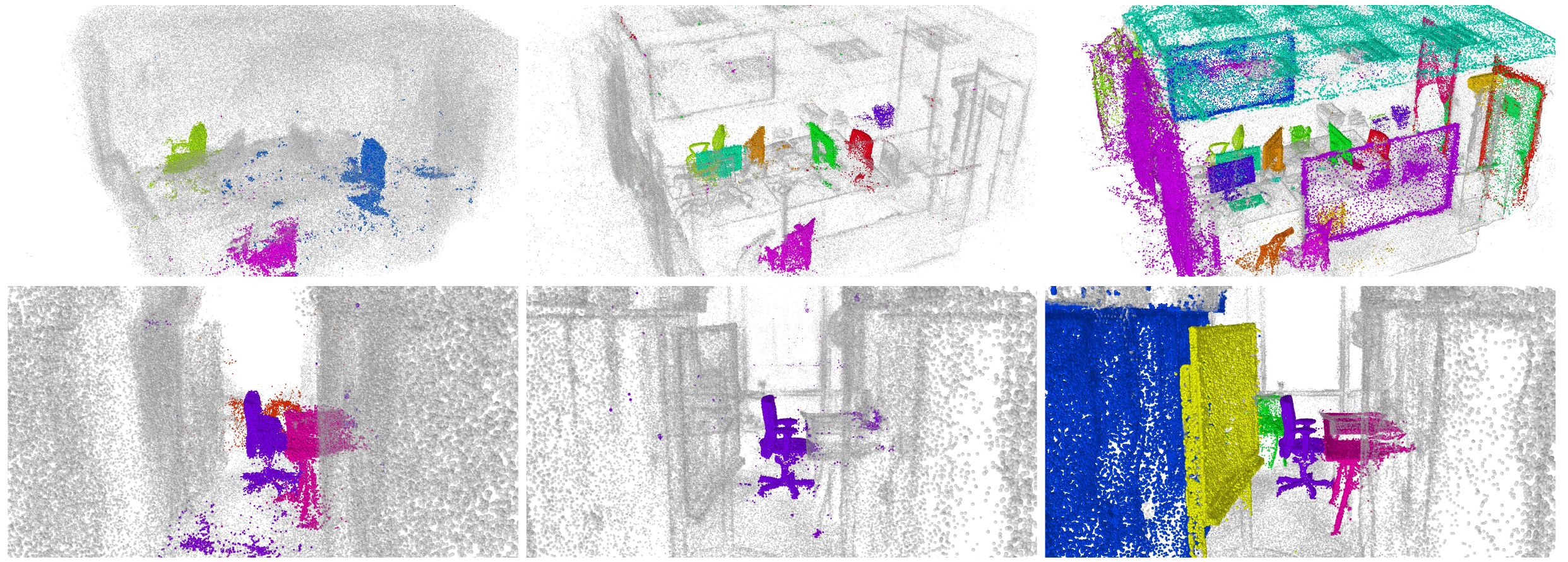}};

        \begin{scope}[
            x={(img.south east)},
            y={(img.north west)},
            every node/.style={anchor=north, font=\small}
        ]
            \node at (0.167, 1.081) {VGGT-$\Omega$};
            \node at (0.500, 1.081) {RaDe-GS};
            \node at (0.844, 1.081) {\textbf{Ours}};
        \end{scope}
    \end{tikzpicture}
    \caption{\textbf{Qualitative results of 3D semantic instance segmentation.} We apply Mask3D directly to colored point clouds reconstructed by each method on ScanNet++. \method{} produces more coherent object masks in the examples shown.
    }
    \label{fig:scannet_seg}
\end{figure}

\paragraph{Point-cloud segmentation.}
We evaluate 3D semantic instance segmentation on ScanNet++ using point clouds extracted from Gaussian splatting, COLMAP, and VGGT-$\Omega$ reconstructions. We feed each method's colored point cloud into Mask3D~\citep{schult2023mask3d} and apply the same pipeline to the ground-truth ScanNet++ point cloud to obtain an oracle reference. \autoref{tab:scannet_miou} shows that \method{} achieves the highest mIoU among the evaluated reconstructed point clouds, more than doubling the Top-1 and Top-3 mIoU of the strongest baseline, VGGT-$\Omega$, and narrowing the gap to the oracle. Textureless regions, such as white walls, remain challenging, but \method{} produces more coherent masks with clearer object boundaries in the examples shown in \autoref{fig:scannet_seg}.

\begin{table}[h!]

\small
\centering
\renewcommand{\arraystretch}{1.15}
\begin{tabular}{lcccccc}
  \toprule
             & COLMAP-S  & COLMAP-D  & RaDe-GS & VGGT-$\Omega$ & \textbf{Ours} & Oracle \\ \midrule
  Top-1 mIoU & 0.02\%  & 8.21\% & 7.77\%  & 10.54\%        & \textbf{24.35\%}           & 30.41\%            \\
  Top-3 mIoU & 0.21\%  & 8.20\% & 7.81\%  & 10.58\%        & \textbf{24.84\%}           & 30.56\%            \\ \bottomrule
\end{tabular}%
\caption{\textbf{Semantic instance segmentation on ScanNet++.}
We report Top-1 and Top-3 mIoU ($\uparrow$) using Mask3D on point clouds reconstructed by each method.
} 
\label{tab:scannet_miou}
\end{table}

\begin{figure}[!h]
    \centering
    \begin{tikzpicture}
        \node[anchor=south west, inner sep=0, outer sep=0]
            (img) at (0,0)
            {\includegraphics[width=\linewidth]{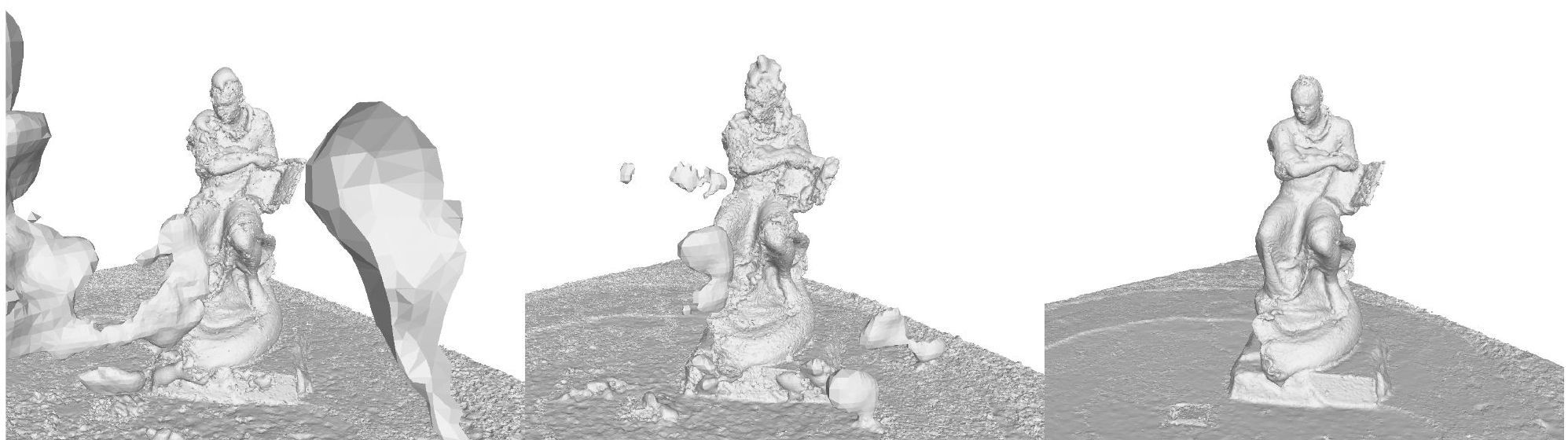}};

        \begin{scope}[
            x={(img.south east)},
            y={(img.north west)},
            every node/.style={anchor=north, font=\small}
        ]
            \node at (0.167, 1.081) {RaDe-GS};
            \node at (0.500, 1.081) {GW};
            \node at (0.844, 1.081) {\textbf{Ours}};
        \end{scope}
    \end{tikzpicture}
    \caption{Using Gaussian centers and normals from \method{}, SPSR recovers finer statue details and a smoother ground surface, with substantially fewer spurious mesh components than with RaDe-GS or GaussianWrapping.
    }
    \label{fig:poisson}
\end{figure}
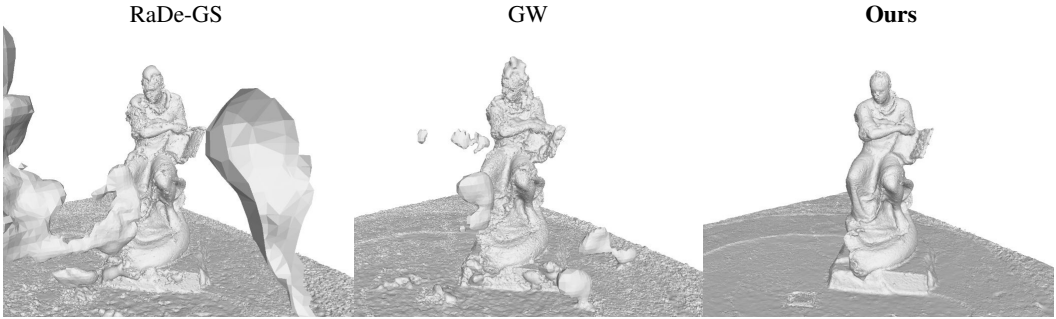

\paragraph{Poisson surface reconstruction.}
To evaluate the quality of the resulting oriented point clouds, defined by Gaussian centers and normals, we apply Screened Poisson Surface Reconstruction (SPSR)~\citep{Kazhdanspsrpoisson}. In the Tanks and Temples example in \autoref{fig:poisson}, \method{} produces cleaner meshes with finer geometric detail and fewer spurious components than RaDe-GS and GaussianWrapping.

%% file: tabs/dtu_pcd_cd.tex
\begin{table}[h]
  \centering
\setlength{\tabcolsep}{2.3pt}
\renewcommand{\arraystretch}{1}
\resizebox{\columnwidth}{!}{%
\begin{tabular}{@{}l*{17}{c}@{}}
  \toprule
  Method & 24 & 37 & 40 & 55 & 63 & 65 & 69 & 83 & 97 & 105 & 106 & 110 & 114 & 118 & 122 & Mean & Time\\
  \midrule
  \hyperlink{cite.wang2026vggtomega}{VGGT-$\Omega$}    &  0.931 & 1.217 & 1.502 & 1.302 & 1.995 & 1.468 & 1.162 & 5.626 & 3.363 & 4.084 & 2.000 & 1.263 & 1.219 & 2.131 & 1.754 & 2.068 & 63s \\
  \hyperlink{cite.schonberger2016structure}{\footnotesize COLMAP-D}         & 0.382 & 0.618 & 0.336 & 0.349 & 0.570 & 0.837 & 0.455 & 1.129 & 0.650 & 0.540 & 0.418 & 0.468 & 0.300 & 0.387 & 0.348 & 0.519 & 26.9m \\
  \midrule
  \hyperlink{cite.kerbl20233d}{3DGS}             & 1.056 & 1.148 & 1.396 & 1.035 & 1.285 & 1.609 & 1.354 & 1.836 & 1.349 & 1.005 & 1.662 & 1.536 & 1.037 & 1.274 & 1.077  & 1.311 & 5.2m\\
  \hyperlink{cite.huang20242d}{2DGS}             & 0.906 & 1.879 & 1.065 & 0.920 & 1.261 & 2.351 & 1.399 & 1.377 & 1.395 & 0.949 & 1.231 & 1.525 & 1.356 & 1.308 & 1.051  & 1.332 & 10.9m\\
  \hyperlink{cite.zhang2024gspull}{GS-Pull}          & 1.243 & 0.803 & 1.127 & 0.719 & 1.129 & 1.698 & 1.554 & 1.744 & 1.493 & 0.954 & 1.891 & 1.791 & 0.806 & 1.395 & 1.107  & 1.297 & 21.8m\\
  \hyperlink{cite.chen2024pgsr}{PGSR}             & \emph{0.621} & \underline{0.651} & 0.817 & \emph{0.611} & \emph{0.804} & \emph{0.780} & \emph{0.729} & 1.020 & \underline{0.800} & 0.647 & 0.801 & \underline{0.651} & \underline{0.510} & 0.666 & 0.571 & \emph{0.712} & 30.5m\\
  \hyperlink{cite.guedon2025milo}{MILo}             & 0.758 & 0.897 & 0.827 & 0.732 & 0.869 & 1.019 & 0.924 & \emph{1.013} & 0.943 & \emph{0.577} & 0.816 & 0.886 & 0.788 & 0.701 & 0.569 & 0.821 & 25.0m\\
  \hyperlink{cite.zhang2026rade}{RaDe-GS}          & 0.776 & \emph{0.764} & \emph{0.790} & 0.656 & \underline{0.732} & \underline{0.648} & \underline{0.614} & \underline{0.956} & \emph{0.857} & \underline{0.505} & \underline{0.496} & \emph{0.696} & \emph{0.558} & \underline{0.433} & \underline{0.445} & \underline{0.662} & 14.5m\\
  \hyperlink{cite.gomez2026blobsspokeshighfidelitysurface}{GW}               & \underline{0.616} & 0.928 & \underline{0.507} & \textbf{0.457} & 0.950 & 0.833 & 0.732 & 1.416 & 0.947 & 0.785 & \emph{0.730} & 1.157 & 0.566 & \emph{0.656} & \emph{0.558} & 0.789 & 18.8m\\
  \textbf{Ours}              & \textbf{0.509} & \textbf{0.551} & \textbf{0.488} & \underline{0.464} & \textbf{0.656} & \textbf{0.563} & \textbf{0.511}  & \textbf{0.910} & \textbf{0.607} & \textbf{0.436} & \textbf{0.425} & \textbf{0.503} & \textbf{0.412} & \textbf{0.371} & \textbf{0.382} & \textbf{0.519} & 22.3m\\
  \bottomrule
  \end{tabular}%
}
 \caption{\textbf{Quantitative comparison on DTU.} We report Chamfer distance ($\downarrow$) between reconstructed and ground-truth point clouds. We highlight the \textbf{best}, \underline{second-best}, and \emph{third-best} Gaussian-based results, with \method{} achieving the best overall results.}
  \label{tab:pcd_dtu}
\end{table}

%% file: tabs/dtu_plus_pca_cd.tex
  \begin{threeparttable}
  \centering
\fontsize{8}{9}\selectfont
\setlength{\tabcolsep}{2pt}
\renewcommand{\arraystretch}{1}
    
    \begin{tabular}{l c @{\;$\rightarrow$\;} c} 
      \toprule
      & \multicolumn{2}{c}{\textbf{Chamfer Dist. ($\downarrow$)}} 
      \\
      \cmidrule(lr){2-3} 
      \textbf{Method} & \textbf{w/o} PCA & \textbf{w/} PCA\\
      \midrule
      2DGS                      & 1.332 & \textbf{1.211}\\
      PGSR                      & 0.712 & \textbf{0.667} \\
      MILo             & 0.821 & \textbf{0.754} \\
      RaDe-GS  & 0.662 & \textbf{0.519} \\
      \bottomrule
    \end{tabular}
    \caption{Effect of local PCA regularization across splatting methods on DTU.}
    \label{tab:other_plus_pca}
  \end{threeparttable}

%% file: tabs/merged_tnt_nerf_f1_pcd.tex
\begin{table*}[h]
  \centering
  \setlength{\tabcolsep}{2pt}
  \renewcommand{\arraystretch}{1}

\resizebox{\columnwidth}{!}{%
  \begin{tabular}{@{}lcccccccc@{\hspace{8pt}}|@{\hspace{8pt}}cccccccccc@{}}
    \toprule
    & \multicolumn{8}{c@{\hspace{8pt}}|@{\hspace{8pt}}}{\textbf{Tanks and Temples}} & \multicolumn{10}{c}{\textbf{NeRF Synthetic Mesh}} \\
    \cmidrule(lr){2-19}
    Method & Barn & Cater. & Court. & Igna. & Meet. & Truck & Mean & Time & Chair & Drums & Ficus & Hotdog & Lego & Mats. & Mic & Ship & Mean & Time \\
    \midrule
    VGGT-$\Omega$\textsuperscript{*} & 0.041 & 0.049 & 0.033 & 0.058 & 0.054 & 0.116 & 0.059 & 2.8  & 0.112 & 0.121 & 0.130 & 0.168 & 0.199 & 0.161 & 0.091 & 0.164 & 0.143 & 1.5\\
    {\footnotesize COLMAP-D}                         & 0.555 & 0.494 & 0.359 & 0.712 & 0.332 & 0.639 & 0.515 & 168.7 & 0.654 & 0.543 & 0.632 & 0.496 & 0.489 & 0.240 & 0.428 & 0.409 & 0.486 & 25.3\\
    \midrule
    3DGS                             & 0.076 & 0.080 & 0.047 & 0.128 & 0.056 & 0.133 & 0.087 & 7.9  & 0.336 & 0.392 & 0.543 & 0.113 & 0.284 & 0.161 & 0.303 & 0.215 & 0.293 & 4.8\\
    2DGS                             & 0.074 & 0.075 & 0.032 & 0.136 & 0.026 & 0.144 & 0.081 & 12.3 & 0.253 & 0.243 & 0.306 & 0.051 & 0.206 & 0.076 & 0.221 & 0.159 & 0.189 & 5.5\\
    GSPull                           & 0.071    & 0.057    & 0.074    & 0.105    & 0.012    & 0.106    & 0.071    & 113     & 0.119 & 0.103 & 0.160 & 0.009 & 0.036 & 0.028 & 0.077 & 0.006 & 0.067 & 26.4\\
    PGSR                             & \emph{0.165} & \emph{0.145} & 0.075 & 0.207 & \emph{0.111} & \emph{0.246} & \emph{0.158} & 72.0 & \emph{0.673} & \underline{0.620} & \textbf{0.874} & 0.313 & \emph{0.462} & \emph{0.338} & \textbf{0.594} & \emph{0.431} & \underline{0.538} & 26.9\\
    MILo                             & \underline{0.324} & \underline{0.258} & \underline{0.172} & \textbf{0.553} & \underline{0.154} & \underline{0.462} & \underline{0.321} & 110  & 0.433 & 0.408 & 0.530 & 0.146 & 0.329 & 0.165 & 0.363 & 0.244 & 0.327 & 25.0\\
    RaDe-GS                          & 0.120 & 0.112 & 0.068 & 0.156 & 0.099 & 0.191 & 0.124 & 18.3 & 0.653 & \emph{0.605} & \emph{0.752} & \underline{0.409} & 0.450 & \underline{0.355} & 0.544 & 0.375 & 0.518 & 10.7\\
    GW                               & 0.153 & 0.134 & \emph{0.079} & \emph{0.223} & 0.088 & 0.234 & 0.152 & 27.0 & \textbf{0.740} & 0.595 & 0.695 & \emph{0.360} & \textbf{0.515} & 0.284 & \emph{0.567} & \underline{0.440} & \emph{0.524} & 13.8\\
    \textbf{Ours}               & \textbf{0.428} & \textbf{0.344} & \textbf{0.231} & \underline{0.509} & \textbf{0.293} & \textbf{0.518} & \textbf{0.387} & 53.0 & \underline{0.727} & \textbf{0.666} & \underline{0.811} & \textbf{0.441} & \underline{0.506} & \textbf{0.386} & \underline{0.593} & \textbf{0.515} & \textbf{0.581} & 25.0\\
    \bottomrule
  \end{tabular}
  \vspace{1ex}
  }
  \raggedright {\tiny \textsuperscript{*} Maximum of 250 images per scene on Tanks and Temples.}
\caption{\textbf{Quantitative comparison on Tanks and Temples and NeRF Synthetic}. We report F1-score ($\uparrow$) between reconstructed and ground-truth point clouds. We highlight the \textbf{best}, \underline{second-best}, and \emph{third-best} Gaussian-based results. Time is in minutes.}
  \label{tab:tnt_nerf_f1_pcd}
\end{table*}

%% file: sections/06_conclusion.tex
\section{Conclusion}

We presented \method{}, a geometry-aware Gaussian Splatting framework that aligns Gaussians to the surface by integrating local PCA regularization into training. Our method achieves surface-aligned Gaussians, which enables downstream tasks while still performing competitively in Novel View Synthesis and Mesh Extraction. This refined representation helps bridge classical point-set surface analysis and modern novel view synthesis, enabling downstream geometry-processing tasks such as 3D segmentation and Poisson surface reconstruction to be applied directly to the optimized Gaussian means. By encouraging Gaussians to lie on the underlying surface, \method{} also provides a natural basis for estimating local geometric quantities, such as curvature and density, opening promising directions for heat propagation~\citep{zhou2025laplace} and geometry-aware pruning and densification (see \autoref{ap:geometry_aware} material). While PCAsplat introduces modest training overhead and requires a depth-based step for distant floaters and sign disambiguation for normals, it substantially improves local surface fidelity

%% file: sections/supp.tex
\newpage

\section*{Appendix}
\section{Implementation details.} \label{sec:implementation_details}

Our method is implemented in \texttt{PyTorch} on top of the RaDe-GS~\citep{zhang2026rade} codebase. We first train the 3D Gaussians with $\mathcal{L}_{\mathrm{RGB}}$ for 7K iterations, and then enable PCA losses, $\mathcal{L}_{\mathrm{PCA}}$ and $\mathcal{L}_{\mathrm{normal}}$, for the remaining iterations. We train bounded scenes for 20K iterations and unbounded scenes for 30K iterations. To accelerate \method{}, we implemented a custom CUDA kernel for batched $3\times3$ eigendecomposition, since \texttt{torch.linalg.eigh} relies on general-purpose routines that are not optimized for this setting. 
All experiments are conducted on an NVIDIA RTX A5500 with 24GB of VRAM.

\paragraph{Results Reproducibility}
We ran all methods using their most up-to-date GitHub repositories on all datasets using their specified hyperparameters for each dataset; however, for NeRF Synthetic we used the DTU hyperparameters for all methods, this is due to the original hyperparameters used for this dataset being PSNR-focused and lacking surface regularizations. For VGGT-$\Omega$, we unproject the predicted depth maps through the predicted camera poses to obtain a metric-scaled point cloud. We then estimate a similarity that aligns the predicted cameras to the dataset COLMAP poses and apply the same transform to the point cloud, placing it in the ScanNet++ coordinate frame. COLMAP-S stands for a sparse point output, while COLMAP-D stands for a dense output. \autoref{tab:pcd_dtu} and \autoref{tab:tnt_nerf_f1_pcd} have all results generated with our runs of each method with their specified hyperparameters for each dataset.

\section{Synthetic Mesh Dataset} \label{sec:nerf_syn_dataset}
\begin{figure}[h]
    \centering
    \includegraphics[width=0.7\linewidth]{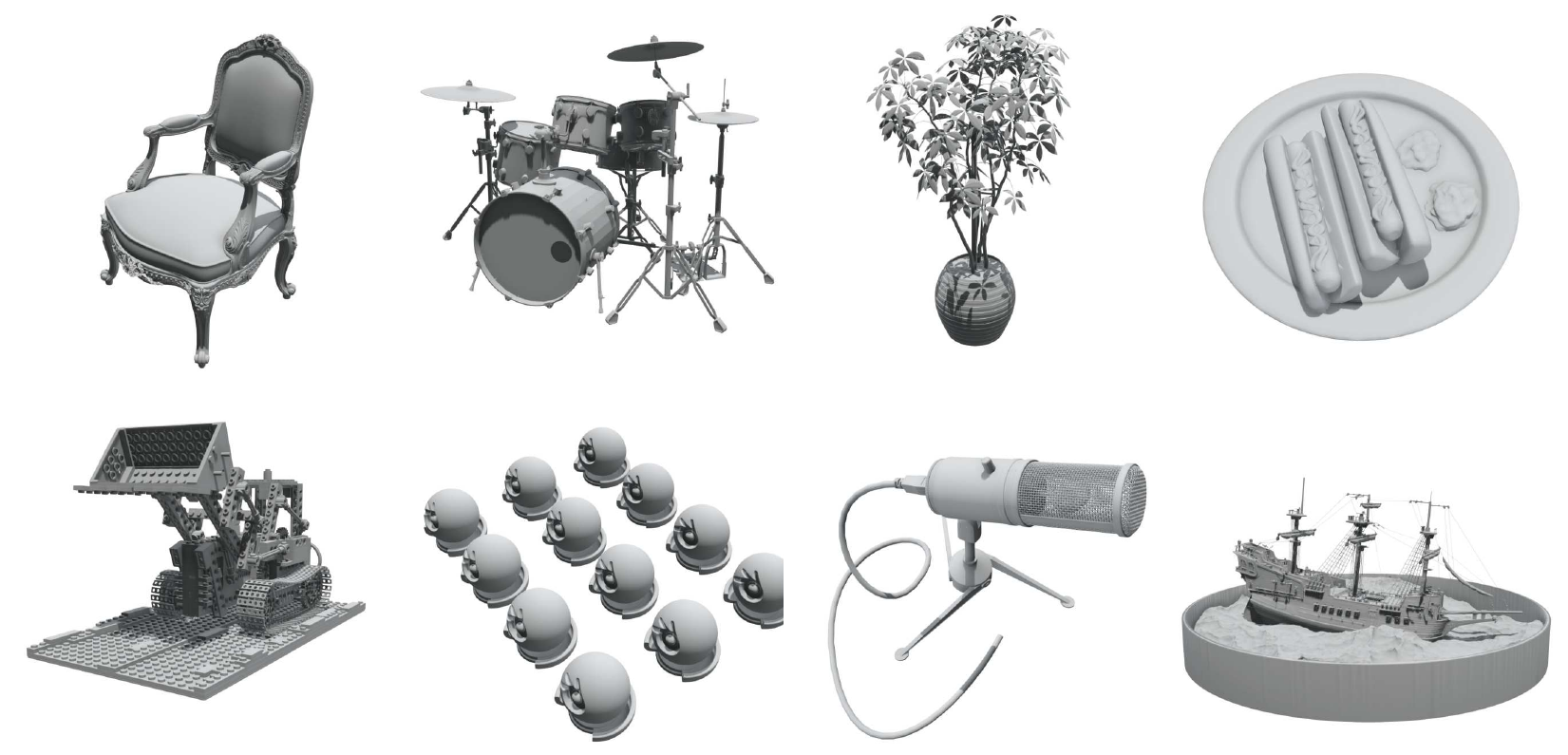}
    \caption{Visible meshes from the NeRF Synthetic eight scenes.}
    \label{fig:ns_meshes}
\end{figure}

NeRF Synthetic~\citep{mildenhall2021nerf} provides synthetic Blender scenes with posed views but no official surface mesh for 3D reconstruction metrics. We extract only the visible mesh from each scene's original 3D assets downloaded from Blendswap. First, all components are merged via self-intersection so shared seams become explicit edges, then hidden-surface removal is run for every train, validation, and test camera, where we raycast each view and keep a triangle if and only if it is the first hit for at least one pixel, yielding a watertight surface that matches what a 3D reconstruction method can actually observe. The resulting visible meshes are the exclusive reference for the F-score measured from $10^6$ Farthest Point Sampling (FPS) samples. This dataset will be made available. See \autoref{fig:ns_meshes}. Additionally, our pipeline for creating such a dataset is useful for evaluating future and previous works on mesh extraction, and creating newer, more precise datasets.

\section{Aditional Results} \label{ap:additional_results}

\begin{figure}[h]
    \centering
    \begin{tikzpicture}
        \node[anchor=south west, inner sep=0, outer sep=0]
            (img) at (0,0)
            {\includegraphics[width=0.8\linewidth]{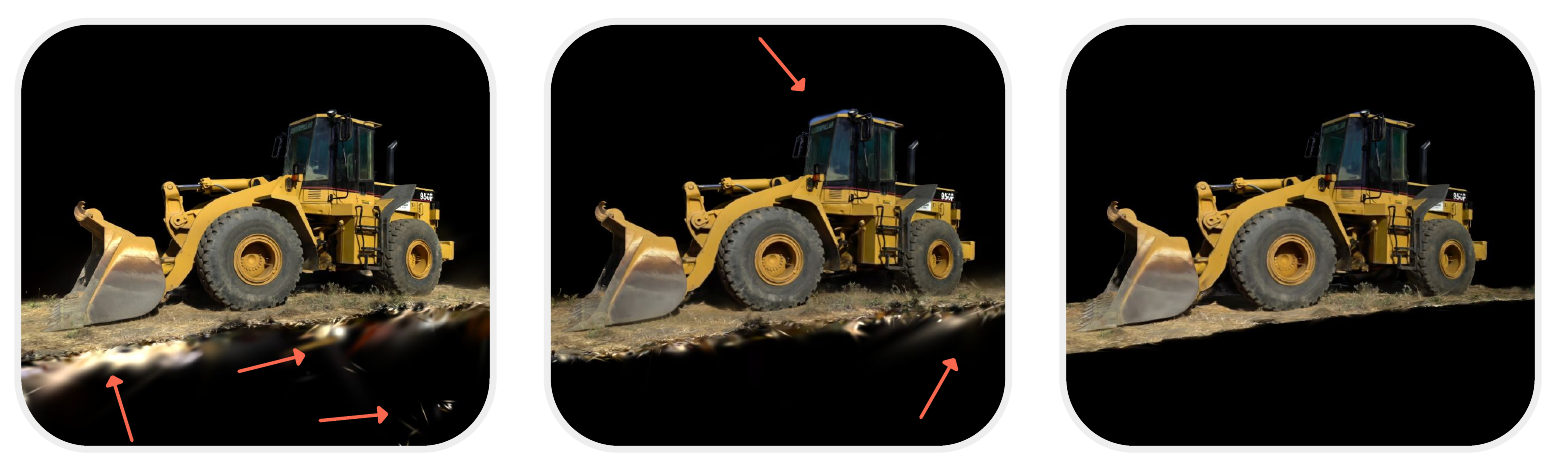}};

        \begin{scope}[
            x={(img.south east)},
            y={(img.north west)},
            every node/.style={anchor=north, font=\small}
        ]
            \node at (0.166, 1.08) {MILo};
            \node at (0.500, 1.08) {GaussianWrapping};
            \node at (0.833, 1.08) {\textbf{\method{} (Ours)}};
        \end{scope}
    \end{tikzpicture}
    \caption{\textbf{Qualitative comparison of object isolation}. Previous methods leave floating primitives and subsurface artifacts (red arrows) upon cropping, needing further manual editing to erase them. In contrast, PCAsplat (Ours) achieves clean, artifact-free boundaries, yielding an asset-ready reconstruction without manual cleanup.}
    \label{fig:asset_ready}
\end{figure}
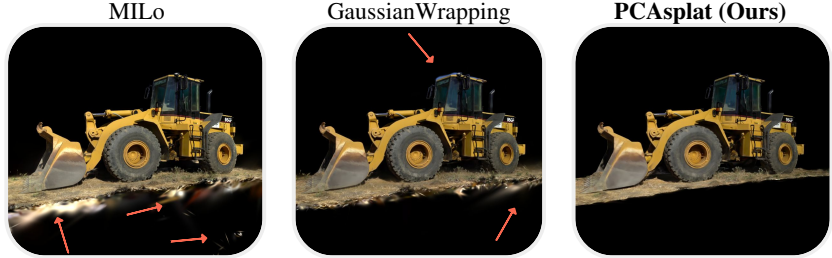

\paragraph{Asset-Ready Splats} Unlike standard 3DGS pipelines, \method{} directly generate asset-ready splats suitable for downstream pipelines. In conventional workflows, isolating a reconstructed object within interactive editing software (e.g., SuperSplat) requires extensive manual cleanup to clean floaters and artifacts. Crucially, such manual pruning often removes important primitives: for instance, filtering Gaussians beneath ground may degrade the opacity of surfaces, introducing transparency. By enforcing surface alignment across the Gaussian primitives, \method{} suppresses unconstrained floating artifacts and as result, target objects can be isolated via simple bounding crops and used out of the box, eliminating manual post-processing while strictly preserving opaque surface integrity. \autoref{fig:asset_ready} showcases a comparison between \method{} and previous works that have surface regularization.  

\begin{figure}[H]
    \centering
    \begin{tikzpicture}
        \node[anchor=south west, inner sep=0, outer sep=0]
            (img) at (0,0)
            {\includegraphics[width=0.7\linewidth]{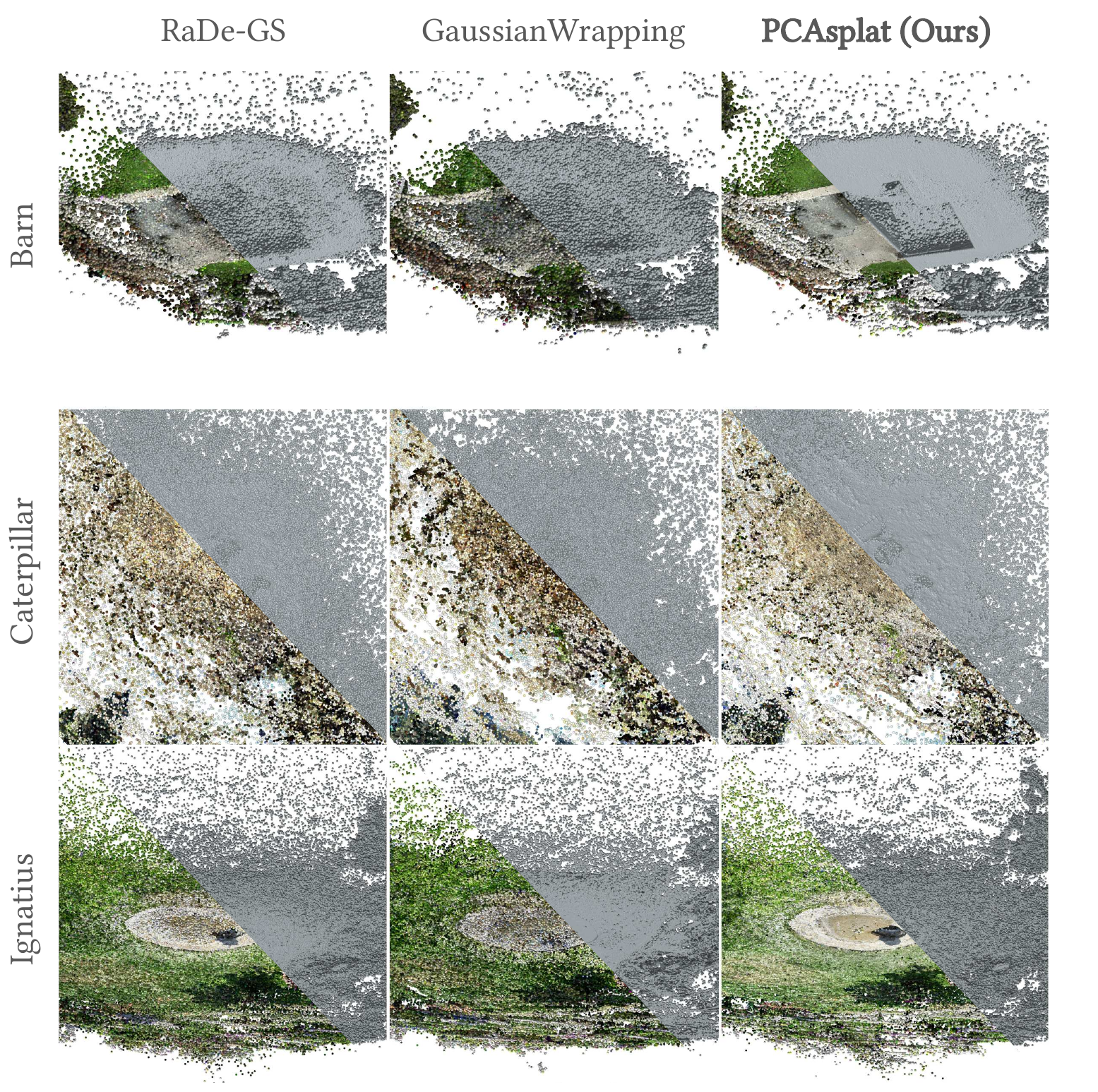}};

        \begin{scope}[
            x={(img.south east)},
            y={(img.north west)},
            every node/.style={
                anchor=base,
                font=\small
            }
        ]
        \end{scope}
    \end{tikzpicture}
    \vspace{-0.3cm}
    \caption{\textbf{``Underworld'' geometry comparison on Tanks and Temples.} We visualize the Gaussian means of reconstructed scenes from beneath the ground. Baselines fail to control interior floaters in unobserved regions and behind structures. \method{} \textbf{(Ours)} suppresses all floaters, resulting in regularized, clean ground and wall boundaries.}
    \label{fig:under_tnt}
\end{figure}

\begin{figure}[h]
    \centering

    \begin{tikzpicture}
        \node[anchor=south west, inner sep=0, outer sep=0]
            (img) at (0,0)
            {\includegraphics[width=0.7\linewidth]{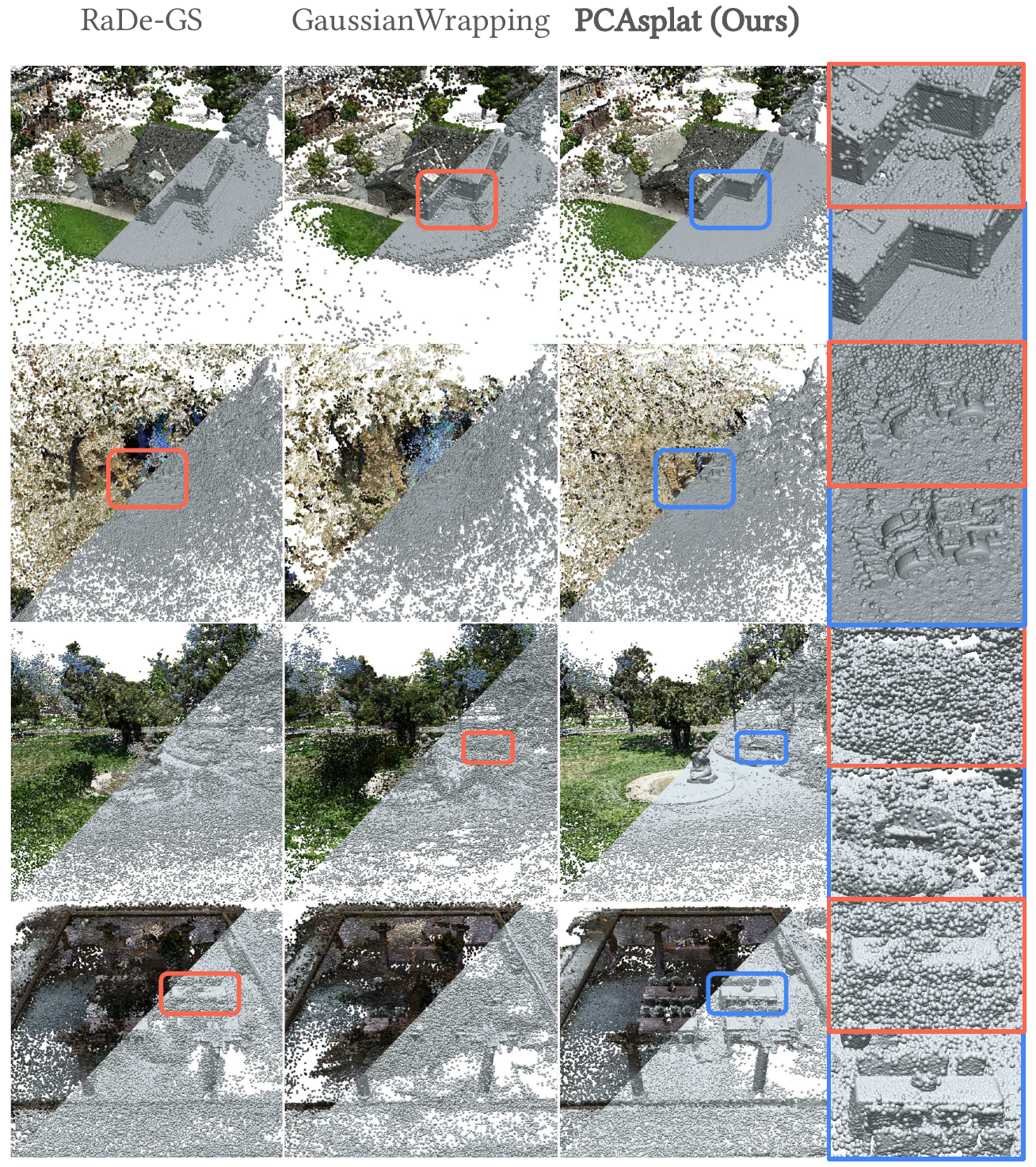}};

        \begin{scope}[
            x={(img.south east)},
            y={(img.north west)},
            every node/.style={
                anchor=base,
                font=\small
            }
        ]
        \end{scope}
    \end{tikzpicture}
    \caption{\textbf{Qualitative point cloud detail comparison on Tanks and Temples} Across diverse outdoor scenes, baseline methods produce noisy clouds. \method{} reconstructs scenes with high geometric fidelity, cleanly resolving intricate structures (e.g., roofs, vehicle contours, and ground details) while removing floating outliers.}
    \label{fig:over_tnt}
\end{figure}

\paragraph{Overworld and underworld} Ray-Contribution supervision alone fails to constrain regions with sparse visibility or with low contributing Gaussians, such as subterranean volumes beneath the ground or occluded regions behind walls. In \autoref{fig:under_tnt}, we show the ``underworld'' from beneath the scene, prior baselines suffer from floaters and disorganized Gaussian sprawl below the ground. In contrast, \method{} regularizes primitives along the floor plane, effectively suppressing subsurface spurious floaters and recovering clean, well-behaved boundary geometry (e.g., the  walls in the \emph{Barn} scene). Concurrently, as shown in the ``overworld'' views in \autoref{fig:over_tnt}, \method{} improves structural contours, accurately recovers thin planar elements, and filters out surface noise across all open world scenes in the Tanks and Temples benchmark.

\subsection*{Surface Reconstruction}
We evaluate \method{} on surface reconstruction using standard mesh extraction from Gaussian splatting representations. Since our method directly regularizes Gaussian centers through local PCA geometry, we also evaluate the learned splat cloud itself as a point-set surface. We report quantitative results on DTU and Tanks and Temples, provide qualitative mesh comparisons, and visualize the spatial distribution of the learned Gaussian point clouds.

\input{tabs/dtu_mesh_cd}

\begin{figure}[H]
    \centering
    \begin{tikzpicture}
        \node[anchor=south west, inner sep=0, outer sep=0]
            (img) at (0,0)
            {\includegraphics[width=0.9\linewidth]{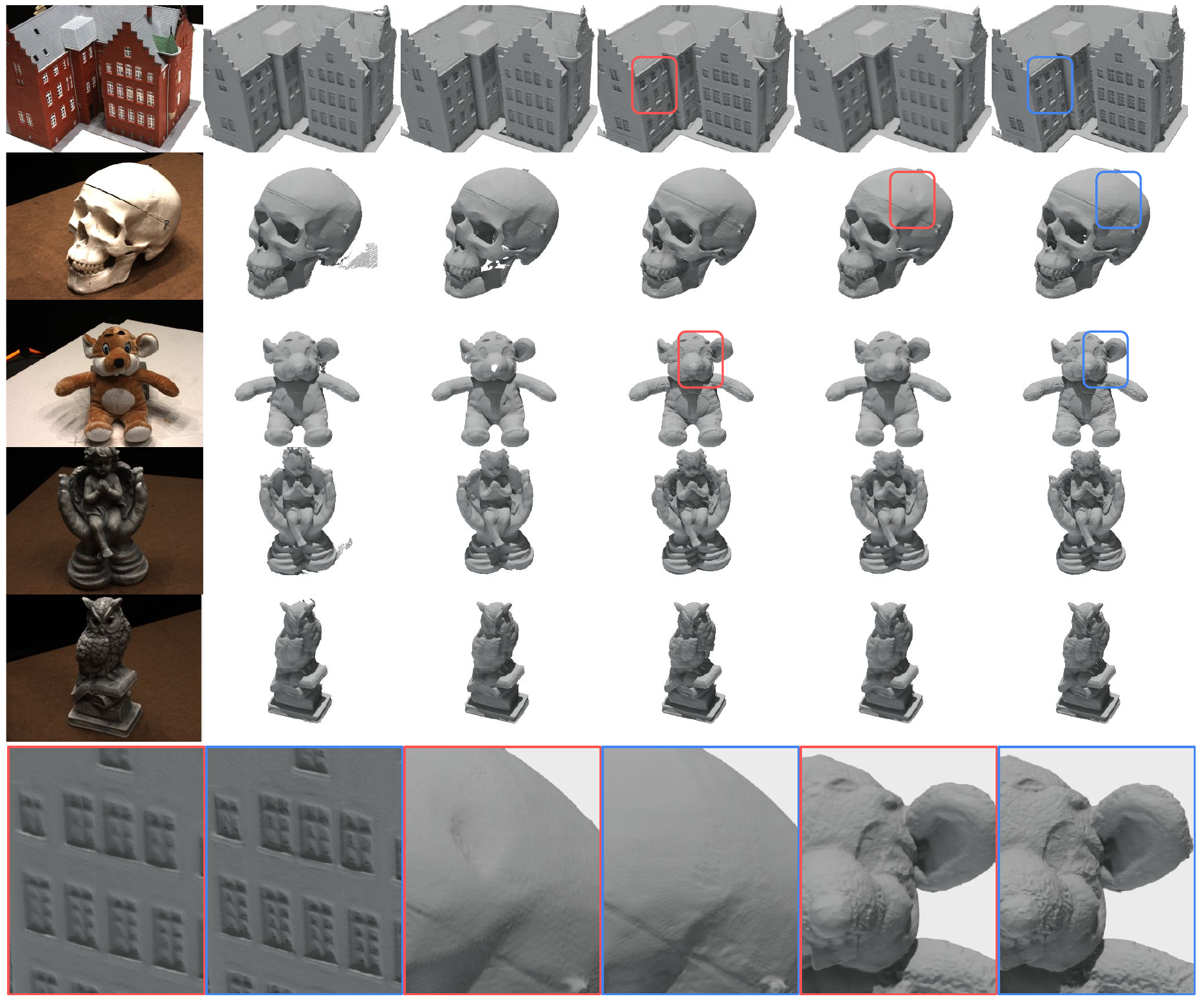}};

        \begin{scope}[
            x={(img.south east)},
            y={(img.north west)},
            every node/.style={anchor=north, font=\small}
        ]
            \node at (0.083, 1.041) {GT};
            \node at (0.250, 1.041) {2DGS};
            \node at (0.417, 1.041) {MILo};
            \node at (0.583, 1.041) {RaDe-GS};
            \node at (0.750, 1.041) {GW};
            \node at (0.917, 1.041) {\textbf{\method{} (Ours)}};
        \end{scope}
    \end{tikzpicture}
    \caption{\textbf{Qualitative comparison on the DTU Dataset.}
    Our method produces cleaner meshes with more details (e.g. teddy bear plush or house windows) and reduced artifacts (e.g. scissor tip, fruit specularity or skull forehead).
    }
    \label{fig:dtu_meshes}
\end{figure}
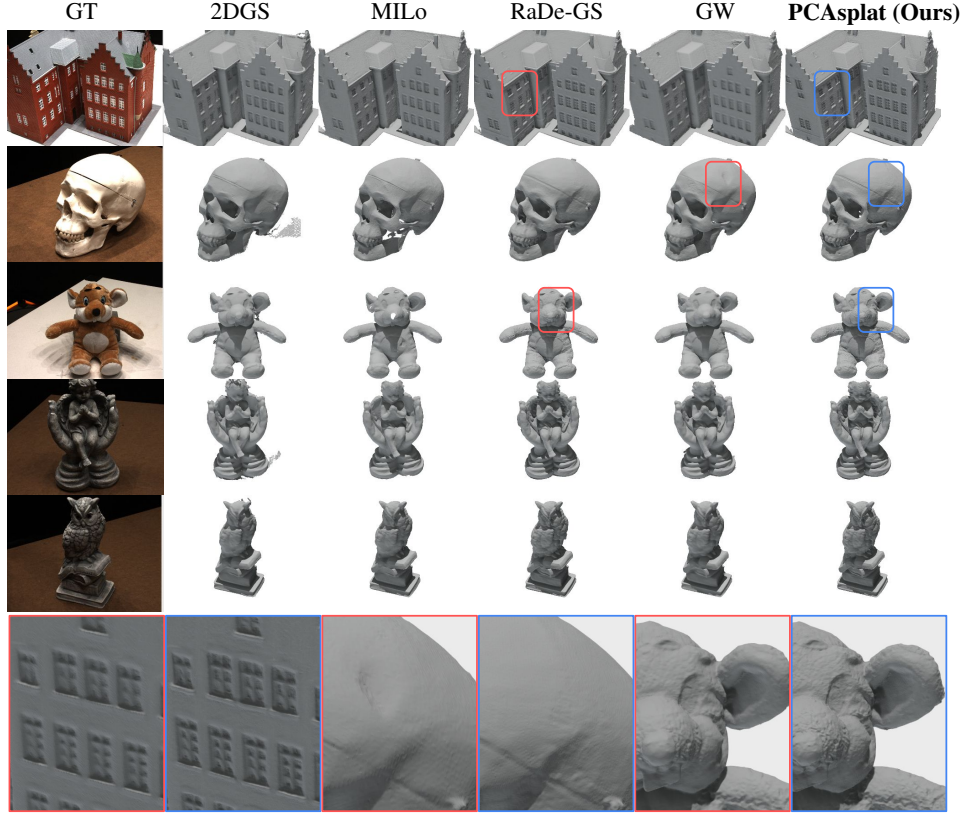

\paragraph{Mesh extraction.}
As shown in \autoref{tab:surface_metrics_dtu} and \autoref{tab:surface_metrics_tant}, \method{} achieves SoTA or competitive performance across both benchmarks. On DTU, our method obtains the lowest mean CD of $0.471$ using TSDF extraction, improving over the strongest baseline. 
On Tanks and Temples, our method achieves a competitive F1-score among Gaussian-based methods, matching the best mean score while maintaining a cleaner underlying point cloud.
Qualitatively, \autoref{fig:dtu_meshes} shows that \method{} produces meshes with finer detail. Compared to 2DGS, RaDe-GS, and MILo, our reconstructions better preserve fine structures, such as building windows, scissor tips, and plush texture, while reducing spurious artifacts around specular and high-curvature regions. Another reason, beyond better geometry, is that our model may have additional learning capacity because its parameters focus on what matters most: the surface.

For evaluation, we followed the standard splatting mesh-extraction pipeline: we render depth maps from all views with a maximum depth of $5$ and fuse them with TSDF~\citep{curless1996volumetric} using a voxel size of $0.002$ using Open3D implemented TSDF.

\input{tabs/tnt_mesh_f1}

\paragraph{Novel View Synthesis}
Our primary focus is improving Gaussian surface alignment; however, to comply with common 3DGS evaluations, we evaluate the novel view synthesis quality of our representation. \autoref{tab:quantitative_results} presents the MipNeRF-360 dataset with our method maintaining competitive performance across all scenes, which demonstrates that our PCA regularization does not hinder the appearance of the scene.

\input{tabs/nvs}

\paragraph{Code integration}

To facilitate integration with existing frameworks, our surface alignment framework is designed as a modular, plug-and-play extension that remains entirely decoupled from the core 3D Gaussian Splatting rasterization pipeline. By encapsulating the Principal Component Analysis (PCA) regularization, users can seamlessly integrate our geometric constraints—such as rotational, isotropic, and surfel losses—into existing optimization loops. As demonstrated in the code snippet \ref{lst:training_loop}, applying our method requires minimal code modification.

\begin{figure}[H]
\begin{lstlisting}[language=Python, caption={Integration of our modular PCA regularization into a standard training loop. The code is entirely decoupled from the core rasterizer.}, label={lst:training_loop}]
pca_pipeline = PCAPipeline(pca_params)
for iteration in range(num_iterations):
    # Accumulate our geometric constraints
    pca_pipeline.update(iteration, gaussians, visibility_filter, opt=opt)
    surface_loss = pca_pipeline.get_surface_loss()
    rotational_loss = pca_pipeline.get_rotational_loss()
    isotropic_loss = pca_pipeline.get_isotropic_loss()
    surfel_loss = pca_pipeline.get_surfel_loss()
\end{lstlisting}
\end{figure}

\section{Other Geometrical Statistics} \label{ap:geometry_aware}

Using our learned PCA eigenspectrum, we leverage well-known formulas for calculating the local density and local curvature of our Gaussian, allowing geometry inspection of trained splats. A shown by \citep{pauly2002efficient} the local curvature for a neighborhood of size $k$ can be evaluated via: 
$$
\sigma_k = \frac{\lambda^{(1)}}{\lambda^{(1)}+\lambda^{(2)} + \lambda^{(3)}}.
$$

Additionally, the point cloud density can be easily calculated via the inverse of the root mean squared distance to its K-nearest neighbors
\begin{equation}
    \rho = \frac{1}{\sqrt{D}} \quad D = \frac{1}{k} \sum_{\mu_j \in N_K(i)} \|\mu_i - \mu_j\|^2.
\end{equation}

\begin{figure*}[h]
    \centering
    \includegraphics[width=0.65\textwidth]{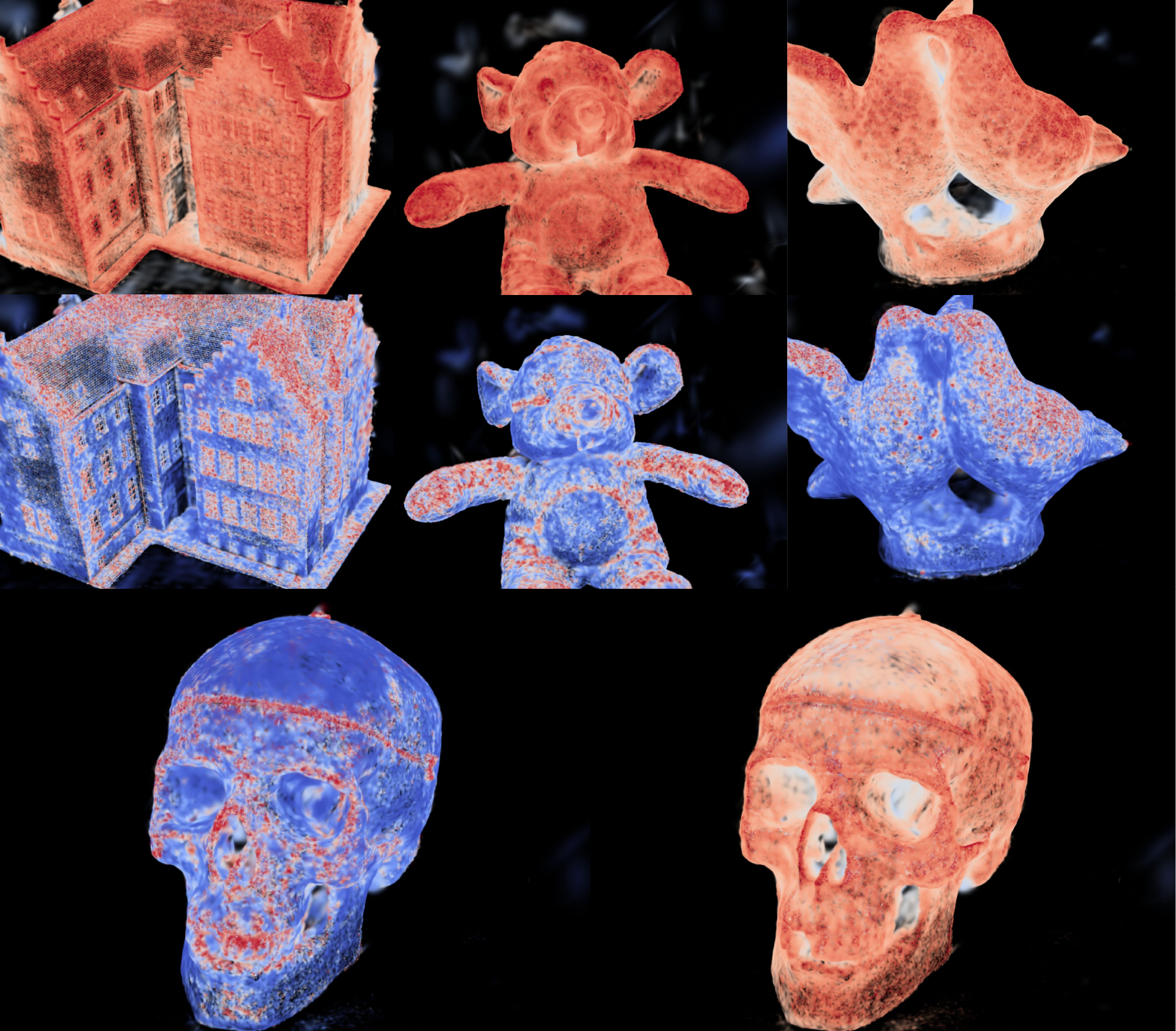}
    \caption{\textbf{Curvature (Blue-Red) and Density (Red) splatting.} Our method enables local curvature and local density rendering.
    }
    \label{fig:curv_dens}
\end{figure*}

We visualize both of these equations applied to the splatting context in \autoref{fig:curv_dens}.



%% file: tabs/dtu_mesh_cd.tex
\begin{table}[H]
\setlength\tabcolsep{0.5em}
\centering
\caption{\textbf{Quantitative comparison on the DTU dataset}. 
We report the Chamfer Distance (CD) $\downarrow$ across 15 scenes and the average optimization time $\downarrow$.
Our method achieves the best mean accuracy with a competitive performance.
}
\resizebox{\textwidth}{!}{%
\begin{tabular}{clccccccccccccccccc}
\hline
\multicolumn{1}{l}{} & \multicolumn{1}{c}{Methods} & 24 & 37 & 40 & 55 & \textbf{63} & 65 & 69 & 83 & 97 & 105 & 106 & 110 & 114 & 118 & 122 & Mean & Time \\ \hline
\multirow{3}{*}{\rotatebox{90}{Neural}} & VolSDF~\citep{yariv2021volume} & 1.14 & 1.26 & 0.81 & 0.49 & 1.25 & 0.7 & 0.72 & 1.29 & 1.18 & 0.7 & 0.66 & 1.08 & 0.42 & 0.61 & 0.55 & 0.857 & 12h> \\
 & Neus~\citep{wang2021neus} & 1.00 & 1.37 & 0.93 & 0.43 & 1.10 & 0.65 & 0.57 & 1.48 & 1.09 & 0.83 & 0.52 & 1.20 & 0.35 & 0.49 & 0.54 & 0.837 & 12h> \\
 & Neuralangelo~\citep{li2023neuralangelo} & 0.37 & 0.72 & 0.35 & 0.35 & 0.87 & 0.54 & 0.53 & 1.29 & 0.97 & 0.73 & 0.47 & 0.74 & 0.32 & 0.41 & 0.43 & 0.610 & 12h> \\ \hline
\multirow{9}{*}{\rotatebox{90}{Gaussian Based}} & 3DGS~\citep{kerbl20233d} & 2.14 & 1.53 & 2.08 & 1.68 & 3.49 & 2.21 & 1.43 & 2.07 & 2.22 & 1.75 & 1.79 & 2.55 & 1.53 & 1.52 & 1.50 & 1.966 & \textbf{5.2m} \\
 & SuGaR~\citep{guedon2024sugar}& 1.47 & 1.33 & 1.13 & 0.61 & 2.25 & 1.71 & 1.15 & 1.63 & 1.62 & 1.07 & 0.79 & 2.45 & 0.98 & 0.88 & 0.79 & 1.330 & 52m \\
 & 2DGS~\citep{huang20242d} & 0.50 & 0.72 & 0.39 & 0.39 & 0.93 & 0.85 & 0.78 & 1.21 & 1.12 & 0.67 & 0.66 & 1.14 & 0.43 & 0.67 & 0.50 & 0.731 & 10.9m \\
 & GS-Pull~\citep{zhang2024gspull} & 0.51 & 0.56 & 0.46 & 0.39 & 0.82 & 0.67 & 0.85 & 1.37 & 1.25 & 0.73 & 0.54 & 1.39 & 0.35 & 0.88 & 0.42 & 0.746 & 21.8m \\
 & GOF~\citep{yu2024gaussian} & 0.48 & 0.82 & 0.39 & 0.38 & 1.36 & 0.86 & 0.75 & 1.20 & 1.31 & 0.83 & 0.65 & 1.30 & 0.48 & 0.69 & 0.53 & 0.802 & 55m \\
 & PGSR~\citep{chen2024pgsr} & \textbf{0.34} & 0.54 & 0.44 & 0.37 & 0.78 & 0.57 & 0.49 & 1.06 & \textbf{0.63} & 0.59 & 0.47 & \textbf{0.50} & \textbf{0.30} & 0.37 & 0.34 & 0.519 & 30.5m \\
 & MILo~\citep{guedon2025milo} & 0.43 & 0.74 & 0.34 & 0.37 & 0.80 & 0.74 & 0.70 & 1.21 & 1.22 & 0.66 & 0.62 & 0.8 & 0.37 & 0.76 & 0.48 & 0.683 & 25.0m \\ 
  & RaDe-GS~\citep{zhang2026rade} & 0.37 & 0.50 & 0.28 & 0.31 & 0.78 & 0.48 & 0.46 & 1.05 & 0.64 & 0.52 & 0.33 & 0.57 & 0.31 & 0.33 & 0.34 & 0.485 & 14.5m \\
  & GW \citep{gomez2026blobsspokeshighfidelitysurface} & 0.40 & 0.51 & 0.27 & 0.32 & 0.81 & 0.46 & 0.45 & 1.04 & 0.65 & 0.52 & 0.33 & 0.59 & 0.32 & 0.33 & 0.34 &0.489 & ---\\\cline{2-19} 
 & \textbf{\method{} (ours)} & 0.36 & \textbf{0.47} & \textbf{0.26} & \textbf{0.30} & \textbf{0.74} & \textbf{0.46} & \textbf{0.44} & \textbf{0.99} & 0.64 & \textbf{0.51} & \textbf{0.32} & 0.63 & 0.31 & \textbf{0.31} & \textbf{0.32} & \textbf{0.471} & 22.3m \\ \hline
\end{tabular}%
}
\label{tab:surface_metrics_dtu}
\vspace{-0.1cm}
\end{table}

%% file: tabs/tnt_mesh_f1.tex
\begin{table}[H]
\centering
\fontsize{8}{9}\selectfont
\setlength{\tabcolsep}{2pt}
\begin{tabular}{clcccccccc}
\hline
& \multicolumn{1}{c}{Methods} & Barn & Caterp. & Court. & Ignat. & Meet. & Truck & Mean & Time \\ \hline
\multirow{3}{*}{\rotatebox{90}{Neural}} & Neus & 0.29 & 0.29 & 0.17 & 0.83 & 0.24 & 0.45 & 0.38 & 12h+ \\
 & Geo-Neus & 0.33 & 0.26 & 0.12 & 0.72 & 0.20 & 0.45 & 0.35 & 12h+\\
 & Neuralangelo & 0.70 & 0.36 & 0.28 & 0.89 & 0.32 & 0.48 & 0.50 & 12h+ \\ \hline
\multirow{9}{*}{\rotatebox{90}{Gaussian Based}} & 3DGS & 0.13 & 0.08 & 0.09 & 0.04 & 0.01 & 0.19 & 0.09 & \textbf{7.9m} \\
 & SuGaR & 0.14 & 0.16 & 0.08 & 0.44 & 0.16 & 0.26 & 0.21 & 73m \\
 & 2DGS & 0.36 & 0.23 & 0.13 & 0.44 & 0.16 & 0.26 & 0.26 & \underline{12.3m} \\
 & GS-Pull & 0.60 & 0.37 & 0.16 & 0.71 & 0.22 & 0.52 & 0.43 & 37.6m \\
 & GOF & 0.51 & 0.38 & 0.26 & 0.68 & 0.24 & 0.52 & 0.43 & 69m \\
 & PGSR & \textit{0.66} & 0.41 & 0.21 & \textbf{0.80} & 0.29 & 0.60 & \textit{0.50} & 72m \\
 & MILo & 0.64 & 0.38 & \underline{0.31} & \textit{0.76} & 0.28 & 0.59 & 0.49 & 110m \\ 
 & RaDe-GS\textsuperscript{*} & \textbf{0.69}\textsuperscript{*} & \underline{0.41}\textsuperscript{*} & 0.18\textsuperscript{*} & \textbf{0.80}\textsuperscript{*} & \textbf{0.40}\textsuperscript{*} & \underline{0.62}\textsuperscript{*} & \underline{0.52}\textsuperscript{*} & \textit{18.3m}\textsuperscript{*} \\
 & GW & \underline{0.65} & \textbf{0.50} & \textbf{0.35} & 0.75 & \underline{0.39} & \textit{0.61} & \textbf{0.54} & 27m \\
 \cline{2-10} 
 & \textbf{\method{}} & 0.64 & 0.38 & \textit{0.21} & 0.73 & \underline{0.39} & \textbf{0.63} & \textit{0.50} & 53m \\ \hline
\end{tabular}
\caption{\textbf{Quantitative comparison on the Tanks and Temples dataset}. 
We report F1-score $\uparrow$.
We evaluate all results using the scores reported in the papers.
\method{} achieves competitive results while producing a cleaner underlying point cloud.
{\tiny \textsuperscript{\textbf{*}} The results presented are from RaDe-GS paper. However, due to compute-limiting constraints, we were unable to re-run RaDe-GS TSDF extraction and achieve the same results as they show in their table. As such, we also append here the average results we got from running their code on our end, \textbf{0.49}, thus, our 0.50 improves over 0.49, and makes our method competitive with SoTA results. }}
\label{tab:surface_metrics_tant}

\end{table}

%% file: tabs/nvs.tex
\begin{table}[H]
\centering

\fontsize{8}{9}\selectfont
\setlength{\tabcolsep}{2pt}
\renewcommand{\arraystretch}{1}
\begin{tabular}{lcccccc}
\toprule
& \multicolumn{3}{c}{Indoor Scenes} & \multicolumn{3}{c}{Outdoor Scenes} \\
\cmidrule(lr){2-4} \cmidrule(lr){5-7}
Method & PSNR $\uparrow$ & SSIM $\uparrow$ & LPIPS $\downarrow$ & PSNR $\uparrow$ & SSIM $\uparrow$ & LPIPS $\downarrow$ \\
\midrule
3DGS            & 30.41          & 0.920          & 0.189          & 24.64          & 0.731          & 0.234          \\
Mip-Splatting   & \textbf{30.90} & 0.921          & 0.194          & 24.65          & 0.729          & 0.245          \\
BakedSDF        & 27.06          & 0.836          & 0.258          & 22.47          & 0.585          & 0.349          \\
SuGaR           & 29.43          & 0.906          & 0.225          & 22.93          & 0.629          & 0.356          \\
2DGS            & 30.40          & 0.916          & 0.195          & 24.34          & 0.717          & 0.246          \\
GOF             & \underline{30.79} & 0.924       & 0.184          & \underline{24.82} & \underline{0.750} & \underline{0.202} \\
RaDe-GS         & 30.74          & \emph{0.928}   & \underline{0.165} & \textbf{25.17} & \textbf{0.764} & \textbf{0.199} \\
MiLo            & \emph{30.76}   & \textbf{0.934} & \textbf{0.155} & \emph{24.81}   & \emph{0.744}   & \emph{0.229}   \\
\midrule
\textbf{Ours }& 30.46          & \underline{0.930} & \emph{0.170} & 24.70         & 0.730          & 0.250          \\
\bottomrule
\end{tabular}%
\caption{Quantitative comparison across indoor and outdoor scenes of the MipNeRF 360 dataset~\citep{barron2022mip}. We highlight the \textbf{best}, \underline{second-best}, and \emph{third-best} PSNR, SSIM, and LPIPS. Our method maintains competitive performance while significantly improving Gaussian localization.} 
\label{tab:quantitative_results}
\end{table}